\documentclass{article}

\usepackage[main, preprint]{neurips_2026}

\usepackage[utf8]{inputenc} 
\usepackage[T1]{fontenc}    
\usepackage{hyperref}       
\usepackage{url}            
\usepackage{booktabs}       
\usepackage{amsfonts}       
\usepackage{nicefrac}       
\usepackage{microtype}      
\usepackage{xcolor}         
\usepackage{amsmath}
\usepackage{graphicx}
\usepackage{natbib}
\usepackage{algorithm}
\usepackage[noend]{algpseudocode}
\usepackage{comment}
\usepackage{amssymb}
\usepackage{enumitem}

\usepackage{tabularx}
\usepackage{tcolorbox}
\tcbuselibrary{breakable}

\newtcolorbox{promptbox}[1]{
  title={#1},
  breakable,
  colback=gray!3,
  colframe=black!45,
  fonttitle=\bfseries,
  boxrule=0.5pt,
  arc=1mm,
  left=1mm,
  right=1mm,
  top=1mm,
  bottom=1mm
}

\newcommand{\SysName}{\textbf{PatchScout}}

\title{\SysName: Thematic Web Data Collection via Information Foraging}

\author{%
    Michael West \quad Eduard Dragut \\
    Department of Computer and Information Science\\
    Temple University\\
    Philadelphia, PA\\
    \texttt{\{michael.west0004, edragut\}@temple.edu} \\
}

\begin{document}

\maketitle

\begin{abstract}
Researchers often need web corpora that are not answers to a single query, but reusable collections of documents spanning many decentralized sources. We refer to this problem as \emph{thematic web data collection}: assembling semantically relevant documents that share a common theme but are scattered across heterogeneous and structurally disconnected regions of the web. Existing approaches, including traversal-based methods and LLM-driven web agents, are limited by local exploration or short-horizon stopping, preventing comprehensive collection.
We formulate this task as a long-horizon information foraging problem over query-induced semantic patches, where the central challenge is deciding when to continue local exploitation and when to switch to new regions. We propose PatchScout, a multi-agent framework that instantiates a class of patch-switching policies to balance local exploitation and adaptive patch switching under a fixed budget.
Experiments on four live-web topics show that PatchScout substantially improves yield and domain coverage over traversal-based and search-augmented reasoning agent baselines. In an applied social science setting with an incomplete prior collection, PatchScout discovers 102 previously unidentified entities, illustrating its ability to expand existing thematic datasets.
\end{abstract}

\section{Introduction}
\label{intro}
Many scientific and societal investigations require constructing reusable thematic corpora from decentralized web sources, where relevant documents satisfy shared semantic criteria yet remain structurally dispersed across heterogeneous domains \citep{lazer2009computational}. For example, a public health researcher studying pandemic-related communication may need to collect documents from hundreds of municipal health departments, while an education researcher identifying Autism Support Programs (ASPs) must gather information scattered across hundreds of university websites \citep{kusters2023spanish,west2025applying}. The same problem arises when assembling evaluation data for a model capability, collecting scientific sources that report a specific phenomenon, or identifying entities that satisfy a given criteria. In each case, the objective is not answering a single query, but assembling a collection of semantically relevant documents drawn from diverse, decentralized sources.

We refer to this task as \emph{thematic web data collection}.  We define a \emph{theme} as a user-provided specification of the target collection, which functions as an operational relevance criterion that defines for each encountered document whether it belongs in the collection. For example, "identify colleges that host an Autism Support Program" is a theme: the target is not a broad subject area, but a concrete set of institutions satisfying a specified condition. The theme is not submitted directly as a search-engine query, instead it is used to guide the entire collection process. Unlike traditional retrieval, where success is measured by the relevance of a ranked list, thematic collection requires simultaneously optimizing \emph{depth} (maximizing thematically relevant yield within discovered sources) and \emph{breadth} (maximizing coverage across distinct sources) under a limited interaction budget. This dual objective creates a fundamentally different search problem: relevant information is sparse, decentralized, and distributed across semantically related but structurally disconnected regions of the web.

One may attempt to address this problem using existing paradigms for web exploration. Traversal-based approaches, such as focused crawling \citep{chakrabarti1999focused, diligenti2000focused, aggarwal2001intelligent}, prioritize link expansion from a set of seed pages based on topical relevance signals. However, these methods rely on the assumption that relevant content is reachable through graph connectivity from initial seeds, an assumption that breaks in decentralized settings where semantically related sources are structurally disconnected. 
Alternatively, query-driven approaches based on search-augmented reasoning generate adaptive queries to retrieve relevant information \citep{jin2025search, qian2025scent, zheng2025deepresearcher}. These systems build on recent work that equips language models with search, browsing, and web-interaction tools to complete question-answering tasks~\citep{nakano2021webgpt,deng2023mind2web}. They often revisit classical retrieval challenges such as selecting promising entry points, iteratively updating relevance estimates, and allocating limited interaction  \cite{dragut2022deep}. However, they are fundamentally designed for \emph{task-bounded} objectives: their stopping criterion is answer sufficiency, not collection completeness. Once sufficient evidence is gathered to answer a user request or generate a coherent report, exploration terminates. As a result, they do not support the sustained exploration required for constructing comprehensive thematic collections.
These observations highlight a fundamental gap: effective thematic web data collection requires jointly reasoning over local exploitation and global re-entry into new regions of the web under a shared budget constraint.

We argue that thematic web data collection is not a page-ranking or task-completion problem; it is a \emph{budget-constrained, long-horizon information foraging problem}. Information Foraging Theory (IFT) models information seeking as movement among information patches guided by information scent and expected marginal gain \citep{pirolli1999information}. Under this perspective, effective thematic collection poses three interdependent challenges: (1) \emph{patch discovery} -- identifying promising new semantic regions; (2) \emph{patch exploitation} -- efficiently extracting high-yield documents within a discovered region; and (3) \emph{patch switching} -- determining when diminishing local returns justify abandoning the current patch in favor of exploring elsewhere. Therefore, the central problem is not relevance estimation or traversal optimization, but strategic patch-switching under budget constraints.

Motivated by this perspective, we propose PatchScout, a multi-agent framework that formulates thematic web data collection as a sequential decision problem over semantic patches and instantiates patch-switching policies for long-horizon exploration. PatchScout enters patches through adaptive search engine querying, exploits discovered patches through local traversal, and explicitly models when to abandon a patch and switch to a new one. This framing transforms thematic collection from traversal-centric crawling or short-horizon search into a long-horizon foraging problem over disjoint semantic patches. We implement this strategy through a multi-agent framework composed of specialized agents for patch discovery, relevance assessment, patch exploitation, and patch switching.

Our contributions are:
\begin{itemize}[nosep]
    \item We introduce thematic web collection as a distinct problem: given a user-defined theme, the objective is to discover and assemble a distributed set of relevant web resources over a long horizon, rather than answer a single query or remain within one topical neighborhood.
    \item We formulate it as a budget-constrained information-foraging problem, treating the decision to continue exploiting the current semantic patch or re-seed into a new region of the web as the central control problem.
    \item We propose PatchScout, whose Scout--Forager--Critic architecture couples adaptive query generation, within-patch exploitation, and between patch exploration through a family of patch-switching policies.
    \item On four live-web topics, PatchScout improves document yield, domain coverage and collection diversity over classical crawlers and deep-research agents.
\end{itemize}

\section{Related Work}
\label{prev_work}
Building collections of semantically related web documents has been studied in several related contexts, particularly in traversal-based focused crawling and domain-specific web data acquisition \citep{olston2010web, pham2018learning, chen2024comquest, chen2025comcrawler, west2026demo}. Early work explores hybrid strategies that combine graph exploration with query-based discovery. \citealt{qin2004building} consider the construction of scientific digital libraries and identify structural limitations of traversal-only approaches, incorporating periodic query-based discovery via search engines. However, their approach relies on a static set of user-provided queries, limiting adaptability and preventing sustained exploration across dynamically discovered regions of the web. More recently, \citealt{pham2019bootstrapping} introduce DISCO, which combines query-based discovery with local graph exploration and models action selection as a multi-armed bandit based on observed marginal yield \citep{auer2002finite}. Their results demonstrate that interleaving these actions substantially improves over traversal-only baselines.

While these approaches recognize the need to combine query-based discovery and local exploration, they treat these operations as competing alternatives with isolated rewards. Thus, the decision problem is framed as a greedy selection between actions based on historical yield. This formulation fails to capture their complementary structure: query-based discovery exposes new regions of the information space, while local graph exploration efficiently exploits those regions for high-yield extraction. The relevant policy is therefore not selecting between actions, but optimizing their coordinated composition over time. More fundamentally, prior work does not explicitly formulate the problem as a long-horizon, budget-constrained collection task over decentralized semantic targets.

Information Foraging Theory (IFT) provides a useful abstraction by modeling information-seeking as navigation over an information landscape composed of \textit{information patches}, regions containing relevant information, and \textit{information scent}, signals that estimate the value of nearby content \citep{chi2001using, pirolli2003snif}. In this setting, a query induces a semantic patch by exposing a set of entry points, while local graph exploration exploits the neighborhood reachable from those points. Under this view, query actions perform patch discovery, traversal actions perform patch exploitation, and the central policy decision is when to switch between patches.

Modern search-augmented reasoning systems address related information-seeking problems. \citet{trivedi2023interleaving} interleave retrieval with chain-of-thought steps to answer multi-hop questions. \citet{jiang2023active} make this process adaptive by introducing forward-looking retrieval triggers based on generation uncertainty that issue queries when model confidence falls below a threshold. \citet{asai2023self} uses supervised fine-tuning to train a model to emit retrieval tokens, allowing the model to learn \textit{when} to search, while \citet{jin2025search} achieves this through reinforcement learning. More recent works have extended this reasoning-search paradigm by incorporating live web search, rather than retrieval from static corpora \citep{li2025search, zheng2025deepresearcher, chen2026research}. \citealt{qian2025scent} interpret such systems through the lens of IFT, modeling intermediate reasoning steps and subqueries as evolving expressions of information scent that guide retrieval toward increasingly useful patches. However, these approaches optimize patch switching implicitly as part of a reasoning trajectory toward a single answer. In contrast, our setting requires treating patch switching as an explicit policy decision and optimizing over long-horizon collection objectives across disjoint regions of the web graph.

Taken together, prior work addresses components of this problem, either through traversal-based exploration, query-based discovery, or short-horizon reasoning, but does not provide a unified formulation of thematic web data collection as a long-horizon, budget-constrained decision process. We address this gap by formalizing thematic web data collection as an information foraging problem over query-induced semantic patches.

\section{Problem Definition}
\label{prob_def}
A thematic collection task involves multi-step semantic search. Consider identifying Autism Support Programs (ASPs) across universities \citealp{west2026evaluating, nachman2022brief}. An agent must first discover which university domains are likely to contain relevant programs. However, domain discovery alone is insufficient. Many universities do not offer dedicated ASPs, while others embed autism-specific services within broader student support ecosystems that also encompass motor, intellectual, visual, hearing, or mental health accommodations. As a result, target relevance is both decentralized and highly heterogeneous: autism-related support may appear as a standalone program, a subprogram within a larger support structure, or may be entirely absent despite superficially relevant institutional signals. Consequently, the agent must determine whether relevant support exists, distinguish autism-specific offerings from broader but non-target programs, and efficiently locate the small number of semantically relevant pages containing critical details such as support type, eligibility, pricing, and accommodations within larger, noisy site structures. Thus, thematic web data collection requires solving both inter-domain discovery (finding promising domains) and intra-domain exploitation (extracting sparse target pages within semantically complex domains), making it a hierarchical search problem characterized by domain uncertainty, semantic nesting, and sparse target information.

We model thematic web data collection as an iterative search process in which a system alternates between patch exploitation and patch switching under a limited interaction budget. Given a thematic objective, the system issues queries to identify promising entry points, locally explores the web regions induced by those entry points, evaluates discovered pages for thematic relevance, and accumulates relevant documents into a growing collection.

In this framework, a semantic patch is induced by a search query and consists of the retrieved seed pages together with their locally reachable neighborhood. Exploiting a patch corresponds to local traversal and relevance-guided extraction within that region, while patch switching corresponds to issuing a new query to enter a different semantic region of the web. Because query-based reseeding provides relatively low-cost access to new regions, the central policy problem is determining when the expected marginal value of continued local exploitation falls below the expected value of switching.

We formalize this decision through four components: (i) semantic patches, (ii) realized information gain, (iii) information scent as an estimate of future within-patch utility, and (iv) the reseed policy governing adaptive patch switching.

\subsection{Semantic Patches}
\label{subsec:semantic_patches}
We model the collection process as alternating between \emph{patch discovery} and \emph{patch exploitation}. While patches can be defined in multiple ways, e.g., via semantic clustering or embedding-based neighborhoods, we adopt a query-induced, traversal-based definition that aligns with how agents access the web through search queries and explore it via link traversal.
Patch discovery begins when the system issues a search query $Q$, and the search engine returns the top-$k$ results $S(Q) = \mathrm{SEARCH}_k(Q) = \{s_{1}, \dots, s_{k}\}$. Let $N_d(s)$ denote the set of pages reachable from search result $s \in S(Q)$ by link traversal within depth $d$. The patch is then:
\[
\mathcal{P}_{k,d}(Q) = \bigcup_{s \in S(Q)} N_d(s).
\]

\subsection{Patch Explosion Under Transitive Closure}
\label{subsec:patch_explosion}

A central challenge in patch-based exploration is that local neighborhoods expand rapidly under transitive closure. Consider a patch induced by a query $Q$ and its top-$k$ results. Let $b$ denote the average branching factor (i.e., the number of outgoing links per page). The size of the reachable neighborhood within depth $d$ grows approximately as
$
|\mathcal{P}_{k,d}(Q)| \leq
k \cdot \frac{b^{d+1}-1}{b-1}
= O(kb^d).
$

Even for modest values (e.g., $k=10$, $b=8$, $d=3$), this yields approximately $5{,}850$ pages. Although overlap between neighborhoods may reduce this number in practice, shallow transitive expansion can still produce thousands of candidate pages within a single patch.
We refer to this phenomenon as \textbf{patch explosion}: a combinatorial growth effect in which local exploration quickly becomes computationally and cognitively impractical. Hence, exhaustive within-patch exploration is not only inefficient, but suboptimal under a limited interaction budget. 
This constraint fundamentally shapes the exploration strategy. To avoid over-investing in a single dense region, an effective agent must balance local exploitation with timely patch switching. The core challenge is thus not simply discovering relevant patches, but managing the tradeoff between exploiting the current patch and re-entering the search space to discover new ones.

\subsection{Patch Utility and Re-Seeding Policy}
\label{subsec:patch_utility}

We quantify the utility of exploration through both realized and expected information gain, and use these signals to define a policy for patch switching.
A \emph{Critic} assigns each visited page $x$ a binary relevance label $r(x) \in \{0,1\}$, where $r(x)=1$ indicates a relevant page. Let $\mathcal{R}_t$ denote the set of relevant pages discovered up to time $t$, such that $\mathcal{R}_t = \{x: r(x)=1\}$. The cumulative information gain is:
$
G_t = |\mathcal{R}_t|.
$
Let $G_m(\tau)$ denote the number of relevant pages discovered during the first $\tau$ steps in patch $m$. We define the \emph{patch productivity} as the average gain per step:
$
\rho_m(\tau) = \frac{G_m(\tau)}{\tau}.
$
%

While $\rho_m(\tau)$ captures the realized return from the current patch, effective decision-making also requires estimating the remaining value of continued exploration. To this end, we model \emph{information scent} as a predictive signal over unexplored links.
At time $t$, let $F_t$ denote the frontier of candidate links within the current patch. For each link $u \in F_t$, the Forager assigns an information-scent score representing the probability that following $u$ yields a relevant page:
\[
\sigma_t(u) = \Pr\!\big(r(x_u)=1 \mid u, c(u), \mathcal H_{<t}\big),
\]
where $x_u$ is the page reached by following link $u$, $c(u)$ is the textual context in which the link appears, and $\mathcal H_{<t}$ is the collection history prior to time $t$.

Let $u_{t,1},\dots,u_{t,h}$ denote the top-$h$ frontier links ranked by $\sigma_t(\cdot)$. We define the \emph{patch-level information scent} as the expected future gain rate under $h$ additional exploration steps:
\begin{equation}
\gamma_t(h) = \frac{1}{h}\sum_{j=1}^{h}\sigma_t(u_{t,j}).
\label{eq:patch_scent}
\end{equation}

Together, $\rho_m(\tau)$ and $\gamma_t(h)$ provide complementary signals: $\rho_m(\tau)$ reflects the realized value of past exploration, while $\gamma_t(h)$ estimates the expected return from continued exploitation.
At each step $t$, the agent observes $s_t = (\rho_m(\tau), \gamma_t(h))$ and selects $a_t \in \{\text{stay}, \text{switch}\}$ according to a threshold-based policy that switches when $\text{trigger}(s_t) < \text{baseline}(\mathcal H_{<t})$, where the trigger is given by either $\rho_m(\tau)$ or $\gamma_t(h)$.
We consider three baselines. A fixed threshold $\eta$ provides a static reference. A global baseline is given by the mean productivity of previously completed patches:
\begin{equation}
\bar{\rho}_{<m} = \frac{1}{m-1}\sum_{i=1}^{m-1} \rho_i(T),
\label{eq:global_patch_avg}
\end{equation}
where $\rho_i(T)$ is the final productivity of patch $i$. To allow the baseline to adapt to changes in patch quality over the course of collection, we also define an exponential moving average (EMA):
\begin{equation}
\theta_{m+1} = (1-\beta)\theta_m + \beta \rho_m(T),
\label{eq:ema_update}
\end{equation}
where $\beta \in (0,1]$ controls the update rate and $\theta_1$ is an initial estimate.
The re-seeding policy is then defined as: 
\textit{\text{switch patch if } \text{trigger} < \text{baseline}.} This policy can be viewed as a threshold-based stopping rule \citep{shiryaev2008optimal}, where the agent switches when the estimated marginal utility of the current patch falls below an adaptive reference level derived from prior experience. This general form induces a family of threshold-based policies obtained by combining different trigger signals and baseline estimators:

\begin{table}[h]
\caption{Re-seeding policies derived from different trigger-baseline combinations. Scent conditions use forward-looking estimates, while realized conditions use observed within-patch performance.}
\centering
\small
\begin{tabular}{llp{8.5cm}}
\toprule
\textbf{Name} & \textbf{Condition} & \textbf{Description} \\
\midrule
Threshold & $\rho_m(\tau) < \eta$ & Realized productivity below a fixed topic-dependent threshold. \\[4pt]
Realized-Mean & $\rho_m(\tau) < \bar{\rho}_{<m}$ & Realized productivity below the mean productivity of prior patches. \\[4pt]
Realized-EMA & $\rho_m(\tau) < \theta_m$ & Realized productivity below the EMA of prior patches. \\[4pt]
Scent-Mean & $\gamma_t(h) < \bar{\rho}_{<m}$ & Information scent below the mean productivity of prior patches. \\[4pt]
Scent-EMA & $\gamma_t(h) < \theta_m$ & Information scent below the EMA of prior patches. \\
\bottomrule
\end{tabular}
\label{tab:reseed_conditions}
\end{table}

These variants allow us to evaluate the relative impact of realized versus predictive signals, as well as static versus adaptive baselines, on long-horizon collection performance.
This formulation makes patch switching an explicit decision driven by diminishing returns. The trigger signals capture both realized and expected utility, while the baselines provide adaptive reference points grounded in past exploration. Together, they implement the tradeoff between continued exploitation of the current patch and re-entry into new regions of the web under a limited budget.

We view patch-based exploration as a sequential decision problem under uncertainty \citep{cassandra1994acting, puterman2014markov}. At each step, the agent must decide whether to continue exploiting the current patch or switch to a new one based on partial observations of realized gain and estimates of future utility. This can be interpreted as an instance of optimal stopping or a two-action decision problem, where the agent terminates local exploration when its expected marginal return falls below that of alternative patches.

\section{Methodology}
\label{method}

We present PatchScout, a multi-agent framework for \emph{thematic web data collection}. Guided by the information foraging formulation in Section~\ref{prob_def}, the system organizes collection as an iterative process over \emph{semantic patches}. At a high level, PatchScout alternates between \emph{patch discovery}, where new queries identify promising regions of the web, and \emph{patch exploitation}, where the system traverses links within a patch to collect relevant pages. A dynamic re-seeding policy determines when to abandon low-yield patches and initiate new ones. The objective is to maximize cumulative information gain under a fixed interaction budget.

\subsection{System Overview}
\label{sys_overview}
PatchScout has three cooperating agents: a \emph{Scout}, a \emph{Forager}, and a \emph{Critic}. These agents implement complementary roles in the information-foraging loop and interact through structured state.

\begin{itemize}[nosep]
    \item \textbf{Scout:} Performs \emph{patch discovery} by generating queries that open new semantic patches and adapts query generation based on past patch outcomes.
    \item \textbf{Forager:} Performs \emph{within-patch exploitation} by traversing links in the current patch, ranking candidate outlinks using information scent, and triggering re-seeding when the current patch becomes unproductive.
    \item \textbf{Critic:} Converts visited pages into realized information gain via relevance classification and provides feedback that informs both query generation and link ranking.
\end{itemize}

\begin{algorithm}[t]
\caption{Thematic Web Data Collection Loop}
\label{alg:collection_loop}
\begin{algorithmic}[1]
\Require collection theme $t$, collection budget $B$
\State initialize visited set $V \gets \emptyset$, relevant set $R \gets \emptyset$, patch history $H \gets \emptyset$
\While{$B$ not exhausted}
    \State $q \gets \textsc{Scout}(t, H)$
    \State $S \gets \textsc{Search}_k(q)$
    \State initialize patch $\mathcal{P}$ from $S$
    \While{$\mathcal{P}$ not exhausted and budget remains}
        \State $u \gets \textsc{Forager}(\mathcal{P}, H)$
        \State visit page $u$ and add $u$ to $V$
        \State $r \gets \textsc{Critic}(u, t, H)$
        \If{$r = 1$}
            \State add $u$ to $R$
        \EndIf
        \State update patch state, frontier, and patch productivity
        \If{\textsc{ReSeed}$(\mathcal{P}, H)$}
            \State \textbf{break}
        \EndIf
    \EndWhile
    \State record completed patch in $H$
\EndWhile
\end{algorithmic}
\end{algorithm}

Algorithm~\ref{alg:collection_loop} summarizes the collection process. Starting from a natural-language theme description, the Scout generates a search engine query which induces a semantic patch. The Forager explores this patch via local link traversal, while the Critic evaluates visited pages for relevance. During exploration, the system tracks expected future gain (Eq.~\ref{eq:patch_scent}) and past productivity (Eq.~\ref{eq:global_patch_avg} or Eq.~\ref{eq:ema_update}). When the re-seeding policy is triggered, control returns to the Scout and the process repeats, with the objective of maximizing cumulative information gain under a fixed interaction budget.

\subsection{Agent Implementation}

\textbf{Scout:} 
The Scout takes as input the user-defined theme and a history of previously explored patches, including queries and their realized productivity. It generates a new query that is both distinct from prior queries and likely to lead to relevant content. The output is issued directly to the search engine to initiate a new patch.\\
\textbf{Forager:} 
The Forager manages within-patch exploration through frontier ranking and re-seeding decisions. It maintains a frontier of candidate links annotated with contextual signals such as anchor text, surrounding text, and page metadata. Using recent exploration history and Critic feedback, it assigns a score to each link, determining its priority in the frontier. At each step, the Forager either continues exploration by selecting a new link or triggers re-seeding according to the specified policy.\\
\textbf{Critic:} The Critic performs binary relevance classification on visited pages. Given the page content (e.g., title and body) and the collection theme, it outputs a relevance label and optionally a short rationale. These outputs are incorporated into the system history and guide both query generation and link ranking.

\subsection{Crawling safeguards}
\label{sec:safeguard}
PatchScout is a read-only research prototype that retrieves publicly accessible web pages. It performs no write actions: it does not authenticate, submit forms, post content, execute transactions, bypass access controls or paywalls, or interact with resources requiring credentials. Its only external actions are HTTP GET requests and search queries issued through a commercial search API. To support polite crawling and limit load on external services, PatchScout reads and respects \texttt{robots.txt} whenever it is available, enforces a global concurrency cap of eight requests, and does not revisit a URL within a run. Requests that return access-control or rate-limit responses, such as \texttt{HTTP 403} or \texttt{429}, are not retried. These controls reduce repeated access, respect site-level crawling restrictions, and avoid continuously requesting blocked or unavailable resources. PatchScout also treats retrieved content as untrusted data rather than executable instructions. Page content cannot alter the system's permissions, trigger write actions, or introduce new tools or credentials. Thus, malicious or prompt-injection content may at most affect a relevance judgment, but it cannot cause external actions beyond the bounded, read-only retrieval process.

\section{Experimental Setup}
\label{exp_setup}
We evaluate PatchScout on \emph{thematic web data collection} tasks on the live web. Our goal is to assess whether dynamic re-seeding improves long-horizon collection performance compared to traversal-centric crawlers and task-bounded agents. We compare a focused crawler, a modern search-augmented reasoning agent, and variants of our framework that differ in their re-seeding policies.

\subsection{Tasks and Evaluation Topics}

We evaluate across four topics selected to vary in \emph{density} (number of relevant domains) and \emph{generality} (breadth of the relevance definition). A key challenge in thematic collection is the absence of ground truth for coverage, since the size and location of the target set are typically unknown. For evaluation, we thus select topics with authoritative entity lists, allowing us to estimate coverage. 
Coverage is measured at the \emph{entity level} rather than by exact domain matching. A target entity may expose relevant content across multiple domains or subdomains (e.g., \texttt{pepsico.com} vs.\ \texttt{pepsicojobs.com}), so discovered pages are mapped to entities prior to computing coverage \cite{dong2018normalization}. Details of this verification process are provided in Appendix~\ref{entity-verification}.

\begin{itemize}[nosep]
    \item \textbf{Four Year Universities (US):} Dense ($N=1{,}928$) and broad; any page within a university domain is relevant~\citep{nces_ipeds}.
    
    \item \textbf{Fortune 500 Employment:} Moderately dense ($N=500$) but narrow; relevance is restricted to employment-related pages~\citep{fortune_500}.
    
    \item \textbf{Accredited Law Schools (US):} Sparse ($N=197$) with broad relevance (any page within the domain)~\citep{aba_law_schools}.
    
    \item \textbf{Autism Support Programs (ASPs):} Extremely sparse ($N=133$) and niche; authoritative list is estimated using datasets from \citet{nachman2022brief} and \citet{viezel2020college}.
\end{itemize}

\subsection{Baselines and Variants}

\textbf{Traversal-centric baseline.}
We use ACHE~\citep{barbosa2007adaptive} to represent traversal-based crawling. ACHE relies on a fixed seed set and expands solely through link traversal without query-based re-entry. We generate seeds using the Scout (query generation) and Critic (filtering). ACHE is run using the public implementation with default settings.\\
\textbf{Task-bounded agent.}
We include DeepResearcher~\citep{zheng2025deepresearcher}, an open-source search-augmented reasoning agent trained for open-domain question answering, which we run locally and can fully inspect. We also evaluate two commercial deep-research systems, Claude Deep Research which is accessible only through the web interface and Gemini Deep Research accessed through API. These closed systems do not expose intermediate reasoning, so we compare them based on their final reports. All three systems receive the same theme specification and prompts which attempt to induce long exploration (Appendix~\ref{deepresearcher-detail}), have no page budget and terminate when they produce a report.  \\
\textbf{PatchScout variants.}
To isolate the impact of re-seeding, we evaluate multiple variants of our framework, which we interpret as instances of a broader policy class over patch-switching decisions: \emph{No-Reseed}, which disables patch switching and performs only initial query-based entry followed by traversal; \emph{Fixed-Reseed}, which triggers patch switching after a fixed number of pages per patch; and the five re-seeding policies of Table~\ref{tab:reseed_conditions} (Threshold, Realized-Mean, Realized-EMA, Scent-Mean, Scent-EMA), each evaluated as a separate system in Table~\ref{tab:main_results} under the same name. Additionally, we conduct component-level ablations, reported in Appendix~\ref{ablation}, to isolate the effects of policy-guided re-seeding, feedback-driven Scout query generation, and Forager link ranking.

\subsection{Implementation Details}

All agents (Scout, Forager, Critic) are implemented using DeepSeek-V3.1 via API. Queries are issued to Google using SearchAPI. Each method is provided the same theme description and prompt configuration, listed in Appendix~\ref{prompts}. 
All systems operate under a fixed budget of 10,000 page visits per run. The budget controls page fetches rather than model compute; token usage for all systems is reported in Section~\ref{results} and Appendix~\ref{cost-analysis}. For consistency, all collected pages from baseline methods are evaluated post-hoc using our Critic Agent. All experiments follow the crawling safeguards specified in Section~\ref{sec:safeguard}.

Our framework introduces a small number of policy-specific hyperparameters. For \textit{Fixed-Reseed}, we use a patch budget of 50 pages. \textit{Threshold} uses a topic-specific value $\eta$, selected based on stable performance observed in preliminary runs.
The \textit{Scent} and \textit{EMA} policies introduce $h$ (frontier size) and $\beta$ (smoothing rate). We perform a small sensitivity analysis on representative dense (Universities) and sparse (ASP) topics (see Appendix~\ref{hyperparam}). We find that $h=10$ and $\beta=0.3$ yield stable performance across topics, and use these values in all experiments.

\section{Results}
\label{results}
\begin{table}[t]
\centering
\caption{Harvest rate (HR\%) and gold-standard domain coverage (Cov\%) at a 10,000-page budget. Deep-research systems are evaluated at completion by coverage over final and intermediate URLs; HR is omitted for them because they have no comparable page-budget denominator. Best completed result per topic per metric in bold.}
\label{tab:main_results}
\small
\setlength{\tabcolsep}{5pt}
\begin{tabular}{lrrrrrrrr}
\toprule
& \multicolumn{2}{c}{\textbf{Universities}} & \multicolumn{2}{c}{\textbf{Fortune 500}} & \multicolumn{2}{c}{\textbf{Law Schools}} & \multicolumn{2}{c}{\textbf{ASPs}} \\
\cmidrule(lr){2-3}\cmidrule(lr){4-5}\cmidrule(lr){6-7}\cmidrule(lr){8-9}
\textbf{System} & HR & Cov & HR & Cov & HR & Cov & HR & Cov \\
\midrule
ACHE & 18.3 & 3.6 & 4.0 & 2.6 & 6.0 & 6.6 & 0.0 & 0.0 \\
DeepResearcher & -- & 0.3 & -- & 0.8 & -- & 1.0 & -- & 17.3 \\
Claude Deep Research & -- & 1.1 & -- & 15.0 & -- & 91.9 & -- & 17.3 \\
Gemini Deep Research & -- & 1.8 & -- & 3.0 & -- & \textbf{93.4} & -- & 9.8 \\
\midrule
No-Reseed & 18.7 & 1.7 & 0.8 & 3.0 & 0.1 & 1.0 & 0.1 & 1.5 \\
Fixed-Reseed & 62.3 & 16.1 & 15.3 & 18.4 & 26.8 & 72.6 & 4.1 & 30.8 \\
\midrule
Threshold & 52.6 & 14.8 & 22.5 & 23.8 & \textbf{38.1} & 85.3 & \textbf{6.1} & \textbf{39.1} \\
Realized-Mean & 73.1 & 2.4 & \textbf{30.1} & 17.0 & 29.5 & 72.6 & 4.9 & 31.6 \\
Realized-EMA & \textbf{86.2} & 4.4 & 23.2 & 19.8 & 25.3 & 70.6 & 3.7 & 26.3 \\
Scent-Mean & 77.7 & \textbf{20.3} & 17.5 & 20.6 & 33.2 & 79.2 & 4.8 & 33.1 \\
Scent-EMA & 68.3 & 15.5 & 20.1 & \textbf{24.6} & 37.2 & 86.8 & 5.6 & 35.3 \\
\bottomrule
\end{tabular}
\vspace{-1em}
\end{table}

\begin{figure}[t]
\centering
\includegraphics[width=\linewidth]{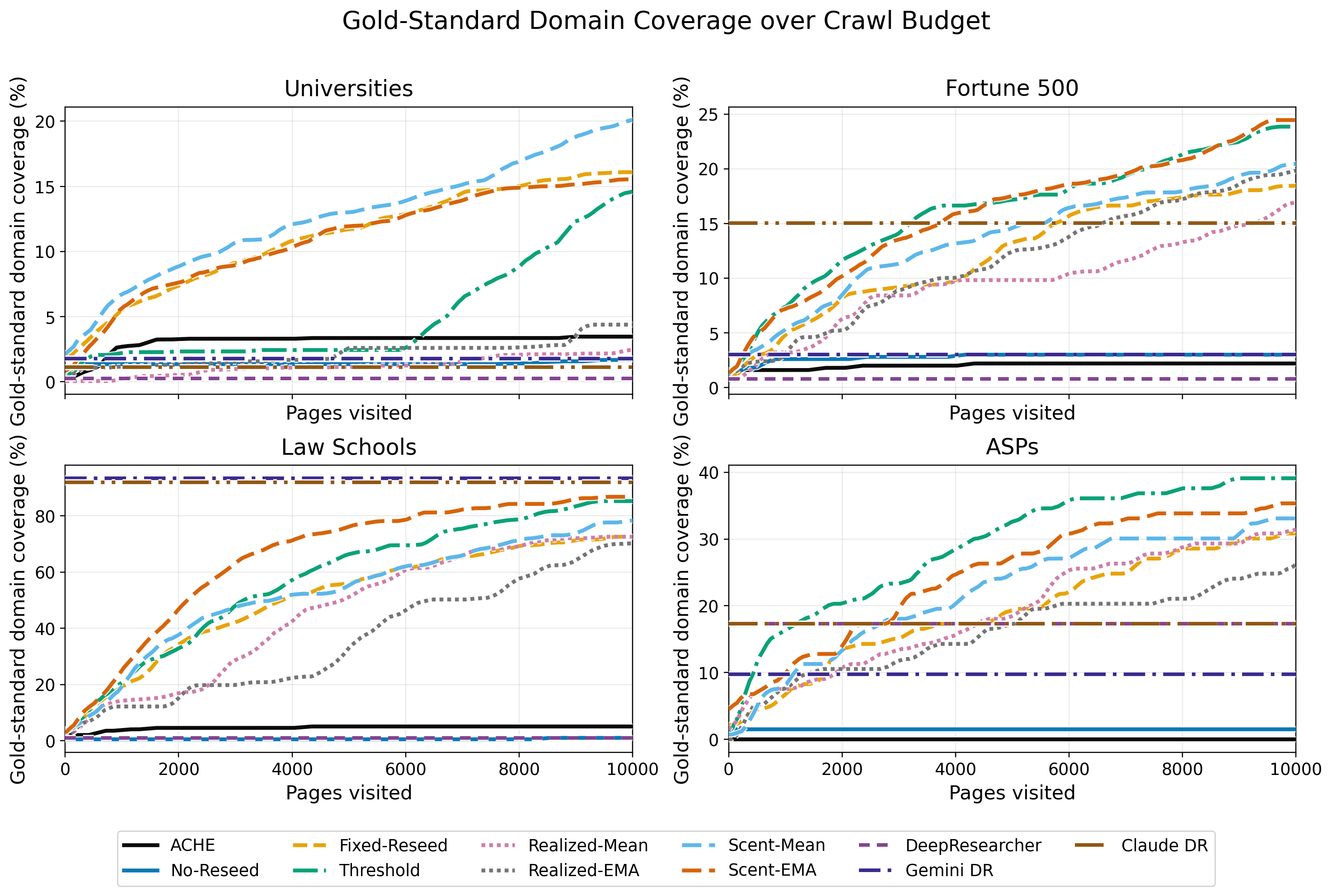}
\caption{Gold-standard domain coverage (\%) over pages visited. Deep-research baselines reported as completion endpoints. Traversal-only methods plateau early, while re-seeding sustains discovery, particularly on sparse topics.}
\label{fig:coverage_curve}
\vspace{-10pt}
\end{figure}

We evaluate all systems on two main metrics: \emph{harvest rate} (the fraction of visited pages classified as relevant) and \emph{domain coverage} (the fraction of ground-truth domains discovered). Table~\ref{tab:main_results} reports endpoint performance after a fixed 10{,}000-page budget. Figure~\ref{fig:coverage_curve} shows coverage growth over time.

\paragraph{Baseline Performance} Traversal-only methods (\textit{ACHE} and \textit{No-Reseed}) plateau rapidly across all topics, especially for sparse settings where weak connectivity limits discovery beyond initial seeds. As shown in Figure~\ref{fig:coverage_curve}, both methods saturate within the first few hundred pages, while re-seeding variants sustain coverage growth throughout collection.
The task-bounded agent baselines do not dominate, but they outperform our best variant on one topic. Law Schools has a small target set ($N=197$) enumerated by a single authoritative accreditation directory: an agent that locates that page can recover nearly the full set without sustained exploration (Gemini 93.4\%, Claude 91.9\%, best PatchScout 86.8\%). When PatchScout reaches the same directory, its roughly 200 outlinks compete in a shared frontier with other , rather than recognizing the list-structure and extracting the record directly \cite{chen2022web}. Where no such source exists, the ordering reverses: PatchScout reaches $11.3\times$ the coverage of the best deep-research system on Universities (20.3\% vs.\ 1.8\%), whose target set is too large for any single directory, $2.3\times$ on ASPs (39.1\% vs.\ 17.3\%), for which no complete list exists, and $1.6\times$ on Fortune 500 (24.6\% vs.\ 15.0\%). This exposes a boundary condition: a theme that is already enumerated reduces to a directory lookup.

\paragraph{Breadth and Depth}
Harvest rate captures local yield, while coverage measures breadth across the target population. These objectives often diverge, reflecting a mismatch between page-level relevance and coverage across distinct domains. For example, \textit{Realized-Mean} achieves the highest harvest rate on Universities (73.1\%) and Fortune 500 (30.1\%), but lags in coverage relative to \textit{Scent-Mean} and \textit{Scent-EMA}. This occurs because productivity-based triggers over-exploit dense patches. On Universities, \textit{Realized-Mean} spends 2{,}457 pages in a single patch (\texttt{uchicago.edu}), yielding 2{,}399 relevant pages but only 28 unique domains, high local yield but poor coverage.

This tradeoff is dependent on topic structure. Dense, broad topics (Universities) favor depth, while sparse or niche topics (ASPs) require aggressive exploration. Fortune 500 Employment lies between these extremes, requiring both domain discovery and within-domain navigation. These results confirm that optimizing for relevance alone can produce narrow, unrepresentative collections.

Scent-based policies generally improve coverage by estimating remaining patch value. \textit{Scent-Mean} achieves the highest coverage on Universities, while \textit{Scent-EMA} performs best on Fortune 500 and Law Schools. However, no policy dominates across all topics. On ASPs, patches are often entirely unproductive, causing adaptive baselines to quickly degrade to near zero values which allows local traversal in unproductive patches. In this case fixed thresholding performs best, suggesting that aggressive switching may outperform fine-grained estimation in highly sparse settings.

For Universities, Fortune 500, and Law Schools, gold-standard lists are treated as closed sets. For ASPs, the list is incomplete and somewhat outdated; we report coverage against the known set and manually verify additional discoveries. PatchScout identifies 102 new ASP domains absent from the prior list, demonstrating its ability to expand existing datasets.

Overall, the best re-seeding policy depends on the target theme and the desired balance between local yield and global breadth. This reinforces the central claim: thematic web-data collection is not solved by relevance estimation alone, but by explicitly modeling patch discovery, patch exploitation, and patch switching under a limited budget.

\paragraph{Cost Analysis} The page budget does not control model compute, so we also compare systems by token consumption (Appendix~\ref{cost-analysis}). Truncating each PatchScout run at the token budget DeepResearcher consumed on the same topic, the best PatchScout variant matches or exceeds DeepResearcher on every topic (Table~\ref{tab:token_matched}). PatchScout is thus more efficient at matched compute and continues converting budget into coverage after task-bounded agents stop, however, these returns decline as the accessible target set is exhausted.

\paragraph{Ablations and Robustness} Ablation studies show that each component of the PatchScout framework contributes to the systems success (Table~\ref{tab:ablation_results}). Random patch switching reduces Law Schools coverage from 86.8\% to 25.4\% and ASP coverage from 35.3\% to 9.0\%; generating all queries upfront reduces Fortune 500 coverage from 24.6\% to 6.6\%; and random link selection reduces Universities harvest rate from 86.2\% to 53.5\%. Results do not depend on a large Critic: binary agreement with our production Critic is 86.1\%, 87.4\%, and 86.9\% for 9B, 27B, and 397B reference models (Table~\ref{tab:critic_4way_relabel}).

\section{Conclusion}
\label{conclusion}

We introduced PatchScout and formulated thematic web data collection as a long-horizon information foraging problem over query-induced semantic patches. This framing highlights the need to balance local exploitation with adaptive patch switching under a limited budget. 
Our experiments show that patch-switching policies fundamentally change collection dynamics: traversal-only methods plateau early, while re-seeding continues discovering new domains. We also find that yield and coverage are distinct objectives, and that scent-based policies improve coverage, though no single strategy dominates across topics. In the ASP setting, PatchScout discovered 102 previously unidentified entities, demonstrating its ability to expand incomplete scientific datasets rather than merely recover known targets. 
Overall, thematic web data collection is not just a retrieval or reasoning problem, but a decision problem over exploration. The patch-based foraging perspective provides a foundation for building systems that construct diverse, reusable web corpora.

\section{Limitations}
\label{limitations}

Our study reflects several practical constraints of web-scale collection. Experiments are conducted on the live web, where evolving content and rankings limit exact reproducibility. PatchScout relies on an LLM-based Critic, introducing variability in relevance assessment, and incurs higher per-page cost than traditional classifiers (see Appendix~\ref{cost-analysis} for cost analysis). 
Finally, re-seeding effectiveness depends on the collection objective and topic structure. Policies based on patch productivity capture local yield but may diverge from coverage, which is difficult to estimate without knowing the target set size. Designing strategies that better balance yield and coverage remains an important direction for future work.

\begin{ack}
This work was supported by the U.S. National Science Foundation awards III-2107213, ITE-2333789, and 2026513.
The authors declare no competing interests.
\end{ack}

\bibliographystyle{plainnat}
\bibliography{mybib}


\appendix
\section{Hyperparameters}
\label{hyperparam}
Table~\ref{tab:hyperparams} lists all hyperparameters that may be set for different variants of PatchScout. Table~\ref{tab:scent-dynamic-hyperparameters} shows the results from the hyperparameter sensitivity analysis for $h$ and $\beta$.
\begin{table*}[h]
\centering
\caption{List of hyperparameters for \SysName.}
\label{tab:hyperparams}
\small
\begin{tabular}{p{1.3cm} p{7cm} p{3cm}}
\toprule
\textbf{Symbol} & \textbf{Role} & \textbf{Default Value} \\
\midrule

$k$ & Number of search results retained per query. & $10$  \\

$b$ & Maximum branching factor (number of links to enque in frontier from a single page). & $10$ \\

$\eta$ & Fixed productivity threshold for re-seeding. & Topic dependent. \\

$\rho_0$ & Initial patch productivity estimate. & Topic dependent. \\

$h$ & Number of frontier links used to estimate information scent. & $10$ \\

$\beta$ & Update rate for patch productivity. & $0.3$ \\

\bottomrule
\end{tabular}
\end{table*}

\begin{table}
\centering
\caption{Topic-separated totals for relevant pages and relevant domains across scent-dynamic hyperparameter settings. Balanced columns are computed across both topics: for each topic, values are normalized by that topic's best value across settings (relevant/max relevant, domains/max domains), then averaged equally across topics.}
\label{tab:scent-dynamic-hyperparameters}
\begin{tabular}{rlrrrrll}
\toprule
\multicolumn{2}{c}{} & \multicolumn{2}{c}{\textbf{University}} & \multicolumn{2}{c}{\textbf{ASP}} & \multicolumn{2}{c}{\textbf{Balanced}} \\
\cmidrule(lr){3-4}\cmidrule(lr){5-6}\cmidrule(lr){7-8}
$h$ & $\beta$ & Relevant & Domain & Relevant & Domain & Relevant & Domain \\
\midrule
5 & 0.1 & 743 & 190 & 72 & 36 & 0.884 & 0.764 \\
5 & 0.3 & 813 & 172 & 73 & 50 & 0.930 & 0.866 \\
5 & 0.5 & 868 & 159 & 49 & 30 & 0.807 & 0.638 \\
10 & 0.1 & 777 & 204 & \textbf{78} & \textbf{43} & 0.941 & 0.864 \\
10 & 0.3 & \textbf{880} & \textbf{235} & 74 & 42 & \textbf{0.974} & \textbf{0.920} \\
10 & 0.5 & 729 & 187 & 60 & 38 & 0.799 & 0.778 \\
25 & 0.1 & 838 & 183 & 30 & 19 & 0.668 & 0.579 \\
25 & 0.3 & 618 & 153 & 70 & 42 & 0.800 & 0.746 \\
25 & 0.5 & 849 & 218 & 32 & 19 & 0.688 & 0.654 \\
\bottomrule
\end{tabular}
\end{table}

\section{Entity-Domain Verification}
\label{entity-verification}
Our gold standard lists for each topic contain a single unique root-domain, each representing an entity that belongs to the target theme. For example, Fortune 500 contains \texttt{amazon.com}. For initial coverage estimation, we perform exact root-domain matching between gold standard lists and the set of relevant pages from each collection ($N=77,532$). In this case, both \texttt{amazon.com/jobs} and \texttt{jobs.amazon.com} would match the gold standard list. However, we notice that many entities contained in the list maintain alternate domains, such as \texttt{amazon.jobs} or \texttt{amazonjobs.com}, so we check relevant pages which are not matched to a gold standard domain ($N=20,427$). We strip the root domains to only the stem (i.e. \texttt{amazon}) and compare to every domain in the gold standard list for exact matches and prefix/suffix dependent matches (i.e. \texttt{amazonjobs.com} would become \texttt{amazonjobs} which matches \texttt{amazon}). The domain is marked as a match if it matches exactly one gold standard domain. We utilize an LLM to process the remaining unmatched pages ($N=5,931$). First, we search the text of the page content and title for the gold standard stems, and add them to the candidate match list if they are found. If this finds fewer than $3$ candidates, we fill the remaining slots based on lexical similarity between the domain stem and gold standard list stems. The LLM is then given the domain, example page URLs, corresponding content from those URLS, and the candidate list. It is prompted to choose \texttt{match, no-match or uncertain}, the matching candidate domain if match is chosen, some brief rationale or evidence, and optionally flag the instance for manual review. Instances that are flagged by the LLM for the manual review flag ($N=14$) and those in which the LLM proposes a matching domain which was not in the candidate list ($N=48$) are manually reviewed by researchers. The remaining relevant pages which do not match to a given gold standard entity ($N=2,259$) generally fall into one of three categories: third party directory such as the ABA list of accredited law schools, news or other article content about the theme generally, but not one entity specifically, or similar entity not on the gold standard list such as companies which are part of the Fortune Global 500 or Universities located outside the US.

\section{Computational Cost}
\label{cost-analysis}
\begin{table}[h]
\centering
\caption{Per component cost metrics. Token usage is aggregated across all 28 collection runs included in Table~\ref{tab:main_results}. Latency per call is in seconds.}
\label{tab:token_by_component}
\begin{tabular}{lllll}
\toprule
Component & Total Tokens & Tokens per Page & Total Calls & Latency per Call \\
\midrule
Forager & 198,415,172 & 707.71 & 49,015 & 6.44 \\
Critic & 44,390,930 & 158.33 & 38,834 & 4.68 \\
Scout & 3,798,127 & 13.54 & 1,666 & 4.02 \\
Fetch & 0 & 0.0000 & 344,588 & 0.75 \\
Search & 0 & 0.0000 & 7,571 & 2.01 \\
\bottomrule
\end{tabular}
\end{table}

\begin{table}[h]
\centering
\caption{Token usage per approach, aggregated across all 4 topics included in Table~\ref{tab:main_results}.}
\label{tab:token_by_approach}
\begin{tabular}{lllll}
\toprule
Approach & Forager Tokens & Critic Tokens & Scout Tokens & Total Tokens \\
\midrule
DeepResearcher & - & - & - & 2,258,943\\
No-Reseed & 39,005,430 & 5,442,936 & 2,472 & 44,450,838 \\
Fixed-Reseed & 23,891,382 & 6,762,778 & 563,082 & 31,217,242 \\
Threshold & 21,800,783 & 6,173,678 & 822,458 & 28,796,919 \\
Realized-Mean & 29,978,153 & 6,195,291 & 341,875 & 36,515,319 \\
Realized-EMA & 30,342,823 & 6,054,647 & 259,603 & 36,657,073 \\
Scent-Mean & 25,317,440 & 6,779,714 & 925,704 & 33,022,858 \\
Scent-EMA & 28,079,161 & 6,981,886 & 882,933 & 35,943,980 \\
\bottomrule
\end{tabular}
\end{table}

\begin{table}[h]
\centering
\caption{Full-budget token efficiency per topic. Tok is total LLM tokens; Pg/1K is relevant pages discovered per 1,000 tokens; Cov/1M is the \emph{number} of reference-set entities discovered per million tokens (not a percentage). Claude Deep Research exposes no usage data. Best value per column in bold.}
\label{tab:token_efficiency_full}
\footnotesize
\setlength{\tabcolsep}{2pt}
\resizebox{\textwidth}{!}{%
\begin{tabular}{lrrrrrrrrrrrr}
\toprule
& \multicolumn{3}{c}{\textbf{Universities}} & \multicolumn{3}{c}{\textbf{Fortune 500}} & \multicolumn{3}{c}{\textbf{Law Schools}} & \multicolumn{3}{c}{\textbf{ASPs}} \\
\cmidrule(lr){2-4}\cmidrule(lr){5-7}\cmidrule(lr){8-10}\cmidrule(lr){11-13}
\textbf{System} & Tok & Pg/1K & Cov/1M & Tok & Pg/1K & Cov/1M & Tok & Pg/1K & Cov/1M & Tok & Pg/1K & Cov/1M \\
\midrule
DeepResearcher & 477K & 0.16 & 10.5 & 340K & 0.01 & 11.7 & 680K & 0.00 & 2.9 & 761K & \textbf{0.08} & \textbf{30.2} \\
Claude Deep Research & -- & -- & -- & -- & -- & -- & -- & -- & -- & -- & -- & -- \\
Gemini Deep Research & 707K & 0.16 & 46.7 & 1.1M & 0.08 & 14.1 & 1.8M & 0.14 & \textbf{102.7} & 1.2M & 0.07 & 10.7 \\
\midrule
No-Reseed & 9.9M & 0.19 & 3.1 & 11.6M & 0.01 & 1.3 & 11.6M & 0.00 & 0.2 & 11.4M & 0.00 & 0.2 \\
Fixed-Reseed & 6.7M & 0.93 & 44.4 & 7.6M & 0.20 & 12.0 & 9.1M & 0.30 & 15.7 & 7.8M & 0.05 & 5.3 \\
Threshold & 3.4M & \textbf{1.53} & \textbf{79.3} & 7.9M & 0.29 & \textbf{15.2} & 9.7M & \textbf{0.39} & 17.4 & 7.8M & \textbf{0.08} & 6.6 \\
\midrule
Realized-Mean & 10.4M & 0.71 & 4.3 & 8.9M & \textbf{0.34} & 9.6 & 9.6M & 0.31 & 14.9 & 7.6M & 0.06 & 5.5 \\
Realized-EMA & 10.9M & 0.79 & 7.4 & 7.9M & 0.29 & 12.6 & 10.2M & 0.25 & 13.7 & 7.7M & 0.05 & 4.6 \\
Scent-Mean & 7.3M & 1.07 & 51.4 & 7.4M & 0.24 & 13.9 & 10.4M & 0.32 & 15.0 & 8.0M & 0.06 & 5.5 \\
Scent-EMA & 7.6M & 0.90 & 37.6 & 8.9M & 0.23 & 13.8 & 11.0M & 0.34 & 15.6 & 8.4M & 0.07 & 5.6 \\
\bottomrule
\end{tabular}
}
\end{table}

\begin{table}[h]
\centering
\caption{Token-matched comparison: pages visited (Pg) and gold-standard domain coverage (Cov\%) for each PatchScout system if stopped at the same total LLM token budget the DeepResearcher baseline used for that topic. Claude and Gemini Deep Research rows show final, unmatched results for reference: Gemini reports only total usage and Claude reports none, so neither can be truncated. ACHE has no LLM token cost and is omitted. Best completed Pg/Cov per topic in bold (baseline reference rows excluded from bolding).}
\label{tab:token_matched}
\footnotesize
\setlength{\tabcolsep}{3pt}
\begin{tabular}{lrrrrrrrr}
\toprule
& \multicolumn{2}{c}{\textbf{Universities}} & \multicolumn{2}{c}{\textbf{Fortune 500}} & \multicolumn{2}{c}{\textbf{Law Schools}} & \multicolumn{2}{c}{\textbf{ASPs}} \\
\cmidrule(lr){2-3}\cmidrule(lr){4-5}\cmidrule(lr){6-7}\cmidrule(lr){8-9}
\textbf{System} & Pg & Cov & Pg & Cov & Pg & Cov & Pg & Cov \\
\midrule
DeepResearcher & 78 & 0.3 & 56 & 0.8 & 28 & 1.0 & 64 & 17.3 \\
Claude Deep Research & - & 1.1 & - & 15.0 & - & 91.9 & - & 17.3 \\
Gemini Deep Research & 113 & 1.8 & 82 & 3.0 & 247 & 93.4 & 87 & 9.8 \\

\midrule
No-Reseed & 716 & 1.4 & 893 & 2.6 & 1,556 & 0.5 & 1,063 & 1.5 \\
Fixed-Reseed & 1,072 & 5.7 & 1,405 & 6.0 & 1,995 & 34.0 & 1,568 & 8.3 \\
Threshold & 1,914 & 2.3 & 1,364 & \textbf{9.0} & 1,871 & 31.0 & 1,579 & \textbf{18.8} \\
\midrule

Realized-Mean & 687 & 0.1 & 1,205 & 3.2 & 1,891 & 16.2 & 1,606 & 9.0 \\
Realized-EMA & 645 & 1.1 & 1,375 & 4.6 & 1,780 & 12.2 & 1,597 & 10.5 \\
Scent-Mean & 982 & \textbf{6.7} & 1,477 & 6.6 & 1,756 & 36.0 & 1,529 & 11.3 \\
Scent-EMA & 951 & 5.6 & 1,223 & 7.6 & 1,664 & \textbf{40.1} & 1,491 & 12.8 \\
\bottomrule
\end{tabular}
\end{table}

We report token usage and latency by system component to characterize the computational cost of PatchScout. 

\textbf{Forager} This component accounts for the largest share of LLM usage, consuming 198.4M tokens, or approximately 708 tokens per visited page, with an average latency of 6.44 seconds per call. This reflects the fact that the Forager is invoked throughout local exploration to rank frontier links, and thus processes many more links than the system ultimately visits. Each URL is passed to the Forager with context which includes the anchor text, surrounding text, source page URL, source page title and source page content snippet. The Forager was implemented with a batch based prompt, with a maximum batch size of $16$. Across all $28$ runs, the Forager scored $659,022$ URLs in $49,015$ API calls (batches) for an average batch size of $13.4$. On average, the \textit{Scent} based re-seeding policies scored approximately $24\%$ fewer URLs than the \textit{Realized} variants, indicating that Forager scores are a good early indicator of patch degradation.

\textbf{Critic} This component consumed 44.4M tokens, or approximately 158 tokens per page, with an average latency of 4.68 seconds per call. This cost is incurred because every visited page must be classified for relevance in order to compute realized information gain and update patch productivity. We implement a heuristic content extractor based on HTML structure which prioritizes meaningful page content such as title, body, paragraph and article tags, and avoids noise such as navigation elements and metadata. For each page, the Critic is given the full title and the first $600$ characters of remaining extracted content. The Critic was implemented with a batch based prompt, with a maximum batch size of $12$. Across all it scored $280,000$ visited pages in $38,834$ calls for an average batch size of $7.2$. 

\textbf{Scout} Finally, the Scout is invoked only when entering a new patch, resulting in substantially lower total usage: 3.8M tokens, or approximately 13.5 tokens per visited page, with an average latency of 4.02 seconds per call. The Scout prompt includes recent query performance history and requests a batch of $5$ search engine queries. The relatively small Scout cost indicates that adaptive re-seeding itself is not the dominant source of computation; rather, the main cost comes from making fine-grained semantic decisions during sustained local exploration, which is required of any web interacting system.

Fetching and search engine calls introduce non-LLM latency, averaging 0.75 and 2.01 seconds per call, respectively, but do not contribute to token usage. The discrepancy between $280,000$ total visited pages and $344,588$ fetch calls is indicative of the challenges of operating on the live web; roughly $19\%$ of fetches fail, with the majority due to 403 error (46.2\%) and 404 error (33.9\%), none of which are retried (Section~\ref{sec:safeguard}). 

All experiments were run from a single Linux workstation with an AMD Ryzen 5965WX 24-Core processor and 512 GB RAM running Ubuntu 20.04.5 and Python 3.12.13. LLM calls were made to the DeepSeek-V3.1 API and search queries were issued through SearchAPI over a standard internet connection. Reported latencies are client-observed wall-clock times and include network overhead, provider-side queuing, API response time, and local processing, rather than isolated model inference time. We limited concurrent DeepSeek API requests to at most $4$ in flight at a time, as preliminary experiments showed that larger concurrency significantly increased per call latency. In total, the $28$ runs reported in Table~\ref{tab:main_results} completed in $83.69$ hours for an average of $2.98$ hours per run.

Tables~\ref{tab:token_by_approach}--\ref{tab:token_efficiency_full} compare PatchScout's compute with the deep-research baselines. The gap is more than an order of magnitude over a full run and is concentrated in the Forager, which is invoked throughout local exploration rather than only at patch boundaries. At matched token budgets (Table~\ref{tab:token_matched}), PatchScout visits 8--70$\times$ more pages than DeepResearcher. Over the full budget its tokens-per-entity is higher on Law Schools and ASPs, reflecting continued exploration after task-bounded agents stop. The Critic sensitivity study (Appendix~\ref{app:critic}) suggests that a much smaller Critic model would reduce cost with little change in labels.

\section{Deep-Research Baselines}
\label{deepresearcher-detail}
We evaluate DeepResearcher-7B in an inference-only setting using the model’s native tool-use protocol, without any additional training or modification to the checkpoint. We load the public release model weights from \hyperlink{Hugging Face}{https://huggingface.co/GAIR/DeepResearcher-7b}, which is released under the Apache 2.0 License, run locally on a single NVIDIA A6000 and iteratively parsed model generations as either final answers or structured tool calls. Web search and browsing were handled through the repository’s existing Flask-based handler: search queries were sent to the Serper API, while requested webpages were fetched with the repository’s simple text browser and summarized by a separate OpenAI-compatible reader model. The original work uses \texttt{qwen-plus} for this reader model, but we choose to use DeepSeek-V3.1 to remain consistent with our other experiments. The architecture requires a number of hyperparameters, which we left on default settings: \texttt{search-top-k=10}, \texttt{max-new-tokens=12,000}, \texttt{temperature=0}, \texttt{top-p=0.95}, \texttt{max-turns=12}.  All model outputs, tool calls, tool observations, timing events and token counts were logged for later analysis. We tested a variety of prompting strategies, included below, to try to induce the model to perform longer horizon discovery. Under all prompting strategies the model chose to terminate before utilizing the maximum tool use turns, and performed only shallow exploration. We utilized the highest coverage result from among the prompting strategies for each topic for the main results reported in Table~\ref{tab:main_results}.

We evaluated Claude Deep Research and Gemini Deep Research  API preview, runs on on all four topics. From each final report we extracted all cited URLs, and for Gemini also the intermediate URLs it visited. These URLs were judged by the same Critic and mapped to entities with the same procedure as all other systems (Appendix~\ref{entity-verification}). Claude Deep Research exposes no token or tool-call statistics; Gemini exposes total token consumption only, so neither can be compared at a matched token budget.

\subsection{DeepResearcher Prompts}
\begin{promptbox}{DeepResearcher Collection Oriented Prompt}
You are not answering a question. You are performing thematic web data collection.

Your goal is to discover as many distinct relevant entities and source domains as possible for the following theme:

\begin{verbatim}
<theme-description>
\end{verbatim}

Do not stop after finding a few examples. Continue searching, browsing, and expanding until you have exhausted the available tool-use budget or can no longer find new relevant entities.

For each search/browse step:
\begin{enumerate}
    \item Identify new candidate entities or domains.
    \item Follow promising links to verify whether they are relevant.
    \item Avoid repeatedly visiting the same domain unless it is needed to verify a new entity.
    \item Prefer breadth across distinct entities over depth within a single entity.
    \item Keep track of entities already discovered and actively search for new ones.
\end{enumerate}

Your final answer should be a structured collection report containing:
\begin{itemize}
    \item A list of discovered relevant entities.
    \item The source URL or domain for each entity.
    \item A brief relevance justification.
\end{itemize}
\end{promptbox}

\begin{promptbox}{DeepResearcher Anti-Answer-Sufficiency Prompt}
The task below is not a question-answering task. Do not stop when you have enough information to write a summary.

Your task is to discover a broad collection of relevant entities and source domains for:

\begin{verbatim}
<theme-description>
\end{verbatim}

You should treat each discovered relevant page as evidence for one entity, but the goal is to maximize the number of distinct relevant entities. A good response is not a polished answer; a good response is a large, diverse, verified collection.

Continue searching and browsing even after you can describe the topic. Early stopping is a failure unless additional exploration is unlikely to discover new entities.

During exploration:
\begin{itemize}
    \item Maintain a running list of discovered entities.
    \item Maintain a list of used queries.
    \item Generate new queries that differ meaningfully from earlier ones.
    \item Follow directory pages, institutional pages, and links that may reveal additional entities.
    \item Switch away from sources that only produce duplicate entities.
\end{itemize}

Your final answer should be a structured collection report containing:
\begin{itemize}
    \item A list of discovered relevant entities.
    \item The source URL or domain for each entity.
    \item A brief relevance justification.
\end{itemize}
\end{promptbox}

\begin{promptbox}{DeepResearcher Exploration Policy Prompt}
You are a thematic web data collection agent.

Your task is to build a broad collection of relevant entities and source URLs for the theme below. This is not a question-answering task: do not stop once you can summarize the topic. Continue exploring to maximize the number of distinct relevant entities discovered.

Theme:
\begin{verbatim}
<THEME>
\end{verbatim}

Collection objective:
\begin{itemize}
    \item Maximize distinct relevant entities.
    \item Prefer breadth across domains over depth within one domain.
    \item Verify relevance using source pages whenever possible.
    \item Avoid duplicate entities and duplicate domains unless needed for verification.
    \item Use search results, directory pages, and discovered links as entry points into new regions of the web.
\end{itemize}

Exploration policy:
\begin{enumerate}
    \item Generate a search query likely to discover relevant entities.
    \item Browse promising results and record distinct relevant entities.
    \item Follow links that may reveal additional entities.
    \item When the current source or query stops producing new entities, switch to a meaningfully different query.
    \item Use prior discoveries and failures to choose later queries.
    \item Continue until the tool budget is exhausted or several diverse searches produce no new entities.
\end{enumerate}

Your final answer should be a structured collection report containing:
\begin{itemize}
    \item A list of discovered relevant entities.
    \item The source URL or domain for each entity.
    \item A brief relevance justification.
\end{itemize}
\end{promptbox}

\section{Ablation Study}
\label{ablation}
\begin{table}[h]
\centering
\caption{Ablation results. Each topic reports harvest rate (HR\%) and gold-standard domain coverage (Cov\%), matching the main results table. Best value within each topic/metric is bolded.}
\label{tab:ablation_results}
\small
\setlength{\tabcolsep}{4pt}
\begin{tabular}{lrrrrrrrr}
\toprule
& \multicolumn{2}{c}{\textbf{Universities}} & \multicolumn{2}{c}{\textbf{Fortune 500}} & \multicolumn{2}{c}{\textbf{Law Schools}} & \multicolumn{2}{c}{\textbf{ASPs}} \\
\cmidrule(lr){2-3}\cmidrule(lr){4-5}\cmidrule(lr){6-7}\cmidrule(lr){8-9}
\textbf{System} & HR & Cov & HR & Cov & HR & Cov & HR & Cov \\
\midrule
Random-Reseed & 72.1 & 4.6 & 7.6 & 2.2 & 27.8 & 25.4 & 4.9 & 9.0 \\
Random-Forager & 53.5 & 1.2 & 10.6 & 2.0 & 21.5 & 11.7 & 4.6 & 12.0 \\
Upfront-Scout & 64.3 & 8.3 & 16.9 & 6.6 & 28.6 & 28.9 & 3.7 & 10.5 \\
Realized-EMA & \textbf{86.2} & 4.4 & \textbf{23.2} & 19.8 & 25.3 & 70.6 & 3.7 & 26.3 \\
Scent-EMA & 68.3 & \textbf{15.5} & 20.1 & \textbf{24.6} & \textbf{37.2} & \textbf{86.8} & \textbf{5.6} & \textbf{35.3} \\
\bottomrule
\end{tabular}
\end{table}

We evaluate three ablations to isolate the main decision components of PatchScout. \textit{Random-Reseed} replaces the policy-guided re-seeding decision with random patch switching, while preserving repeated entry into new search-induced patches. Analysis of our re-seeding policy variants revealed that across topics and policies reseeding is triggered on roughly 10\% of decision points, so we enforce a 10\% chance to reseed randomly. \textit{Upfront-Scout} generates all queries before collection begins, similar to prior work on seed finding \citep{vieira2016finding}, removing feedback from observed patch outcomes and Critic judgments during later patch discovery. These two ablations utilize the \textit{Scent-EMA} re-seeding policy, as main experiments find it to have the best consistent performance. \textit{Random-Forager} replaces information-scent based frontier ranking with random link selection, testing the value of guided within-patch exploitation. Given that this ablation does not utilize Forager scores and thus has no \textit{scent} signal, we utilize \textit{Realized-EMA} as the reseeding policy. These ablations respectively test whether performance depends on patch switching policy, adaptive query generation and information-scent based link selection.

As shown in Table~\ref{tab:ablation_results}, all three ablations reduce performance relative to the full PatchScout system. The performance under \textit{Random-Reseed} shows that coverage gains are not explained solely by the number of re-seeding events; rather, the timing of patch switches matters. The weaker performance of \textit{Upfront-Scout} suggests that collection history is informative for the Scout, allowing it to formulate better queries over the course of collection. Finally, the degradation of harvest rate under \textit{Random-Forager} indicates that the Forager agent link priority scores are meaningful predictions of URL relevance, thus guiding local exploitation to high-yield pages and serving as informative information-scent.  Overall, the ablations show that PatchScout's performance depends on the interaction between adaptive discovery, guided exploitation, and policy-based re-seeding.

\begin{table}[h]
\centering
\caption{Pairwise agreement between critic models on the binary relevance decision. For each
pair, agreement (Agr\%), precision (Prec\%), and recall (Rec\%) treat the \emph{row} model as
the prediction and the \emph{column} model as the reference. All four models judged the same 10,000 pages, with the same binary
Yes/No prompt; only the model differs.}
\label{tab:critic_model_agreement}
\footnotesize
\setlength{\tabcolsep}{3pt}
\begin{tabular}{lrrrrrrrrrrrr}
\toprule
& \multicolumn{12}{c}{\textbf{Reference model}} \\
\cmidrule(lr){2-13}
& \multicolumn{3}{c}{\textbf{DeepSeek}} & \multicolumn{3}{c}{\textbf{Qwen 9B}} & \multicolumn{3}{c}{\textbf{Qwen 27B}} & \multicolumn{3}{c}{\textbf{Qwen 397B}} \\
\cmidrule(lr){2-4}\cmidrule(lr){5-7}\cmidrule(lr){8-10}\cmidrule(lr){11-13}
\textbf{Prediction} & Agr & Prec & Rec & Agr & Prec & Rec & Agr & Prec & Rec & Agr & Prec & Rec \\
\midrule
DeepSeek & -- & -- & -- & 86.1 & 81.1 & 86.9 & 87.4 & 79.9 & 90.8 & 86.9 & 77.6 & 91.9 \\
Qwen 9B & 86.1 & 86.9 & 81.1 & -- & -- & -- & 88.7 & 83.6 & 88.6 & 88.0 & 80.9 & 89.4 \\
Qwen 27B & 87.4 & 90.8 & 79.9 & 88.7 & 88.6 & 83.6 & -- & -- & -- & 92.0 & 87.8 & 91.5 \\
Qwen 397B & 86.9 & 91.9 & 77.6 & 88.0 & 89.4 & 80.9 & 92.0 & 91.5 & 87.8 & -- & -- & -- \\
\bottomrule
\end{tabular}
\end{table}

\begin{table}[h]
\centering
\caption{Critic 4-way relabel of sampled pages, reclassified in their original critic batches (same pages, same order, same batch-mates as the first pass): full breakdown of where originally \emph{relevant} and originally \emph{not-relevant} pages land under the four graded labels (not\_relevant / topically\_related\_no\_evidence / partial\_evidence / direct\_evidence). Rows sum to 100\% (rounding). Combined pools all four topics.}
\label{tab:critic_4way_relabel}
\small
\setlength{\tabcolsep}{4pt}
\begin{tabular}{llrrrrr}
\toprule
\textbf{Topic} & \textbf{Orig. label} & \textbf{Not rel.} & \textbf{Topical, no ev.} & \textbf{Partial ev.} & \textbf{Direct ev.} \\
\midrule
Universities & Relevant & 5.8 & 37.9 & 9.3 & 47.1 \\
Universities & Not relevant & 72.9 & 19.5 & 2.0 & 5.6 \\
\midrule
Fortune 500 & Relevant & 5.8 & 9.8 & 36.3 & 48.1 \\
Fortune 500 & Not relevant & 85.4 & 10.3 & 3.8 & 0.4 \\
\midrule
Law Schools & Relevant & 2.8 & 22.4 & 41.8 & 32.9 \\
Law Schools & Not relevant & 66.6 & 28.0 & 5.2 & 0.2 \\
\midrule
ASPs & Relevant & 1.4 & 12.4 & 58.0 & 28.2 \\
ASPs & Not relevant & 77.8 & 17.9 & 4.3 & 0.0 \\
\midrule
Combined & Relevant & \textbf{4.2} & \textbf{21.6} & \textbf{33.8} & \textbf{40.4} \\
Combined & Not relevant & \textbf{75.9} & \textbf{18.8} & \textbf{3.9} & \textbf{1.4} \\
\bottomrule
\end{tabular}
\end{table}

\section{Critic Sensitivity}
\label{app:critic}
\paragraph{Graded relevance.} We relabelled 9,983 pages---4,457 originally judged relevant and 5,526 originally judged not relevant---on a four-point evidence scale (not relevant; topically related without theme evidence; partial evidence; direct evidence). Pages were sampled randomly across all four topics, targeting balance between the two original labels; because sampling operated on whole Critic batches, balance could not be maintained on ASPs, where relevant pages are scarce. Pages were reclassified in their original batches with the same ordering and batch-mates, so batching effects do not confound the comparison. Table~\ref{tab:critic_4way_relabel} shows that the binary boundary is meaningful but loose at the margin: 74.2\% of pages labelled relevant carry at least partial theme evidence and 40.4\% carry direct evidence, while 25.8\% are either not relevant (4.2\%) or topically related without theme evidence (21.6\%). The false-negative rate is low: 1.4\% of pages labelled not relevant carry direct evidence. Universities is the outlier, with 43.7\% of relevant pages lacking direct evidence; this is partly a rubric artefact, since that topic's criterion is any page within a university domain, so topically related pages without specific evidence still satisfy the task definition.
\paragraph{Critic model.} We re-judged 10,000 pages with three additional models using the identical binary prompt (Table~\ref{tab:critic_model_agreement}). Agreement with the production Critic is flat in model scale (86.1\%, 87.4\%, and 86.9\% for 9B, 27B, and 397B), so binary judgment over a title and short snippet appears to saturate well below the scale we use \citep{dietz2025principles}. Within-family agreement (88.0--92.0\%) exceeds cross-family agreement (86.1--87.4\%), so the cross-family figures are the conservative estimate of reproducibility. Disagreement is systematic: treating our Critic as the prediction, its precision against successively larger references falls (81.1, 79.9, 77.6) while recall rises (86.9, 90.8, 91.9), so it labels relevant a superset of what larger models accept. Both analyses indicate that harvest rate is optimistic at the margin; since baseline pages are scored by the same Critic, this affects absolute rates rather than the comparison between systems.

%

\section{Agent Prompts}
\label{prompts}

PatchScout uses three LLM-backed agents. The templates below show the prompts used in the experiments, with run-specific values represented by angle-bracketed placeholders.

\subsection{Scout Agent}

\begin{promptbox}{Scout System Prompt}
You are the Scout Agent in a patch-based thematic web-data collection system. Your job is to propose high-precision search queries that discover new relevant patches. Return only valid JSON. Do not include markdown, prose, or code fences.
\end{promptbox}

\begin{promptbox}{Scout Discovery Prompt}
Task: Generate exactly \texttt{<n\_queries>} distinct search queries for patch discovery.

\medskip
\noindent Theme description:
\texttt{<theme\_description>}

\medskip
\noindent \texttt{<topic-specific rules>}

\medskip
\noindent Requirements:
\begin{enumerate}[leftmargin=1.5em, itemsep=0pt]
  \item Return exactly \texttt{<n\_queries>} executable search queries.
  \item Prefer precision over recall.
  \item Prioritize authoritative, information-dense sources.
  \item Do not produce queries that differ only by term order, pluralization, or synonym swaps.
  \item At most two queries in this batch may start with the same operator prefix (e.g., \texttt{site:*.edu}, \texttt{inurl:...}).
  \item Use exclusion terms selectively; do not repeat long negative suffix bundles across the whole batch.
  \item Prefer HTML-first discovery pages.
  \item Avoid filetype filters by default; do not use \texttt{filetype:pdf}.
  \item Do not output explanations outside JSON.
\end{enumerate}

\medskip
\noindent Return JSON with exactly this schema:
\begin{verbatim}
{
  "queries": ["q1", "q2"],
  "reason": "short rationale"
}
\end{verbatim}
\end{promptbox}

\begin{promptbox}{Scout Re-seed Prompt}
Task: Generate exactly \texttt{<n\_queries>} new re-seed queries for discovering fresh patches.

\medskip
\noindent Theme description:
\texttt{<theme\_description>}

\medskip
\noindent \texttt{<topic-specific rules>}

\medskip
\noindent Here are the most recent queries you've suggested and their outcomes:
\texttt{<recent\_query\_outcomes>}

\medskip
\noindent Requirements:
\begin{enumerate}[leftmargin=1.5em, itemsep=0pt]
  \item Return exactly \texttt{<n\_queries>} new queries.
  \item Queries must be meaningfully different from recently used ones.
  \item Do not produce queries that differ only by term order, pluralization, or synonym swaps.
  \item At most two queries in this batch may start with the same operator prefix (e.g., \texttt{site:*.edu}, \texttt{inurl:...}).
  \item Use exclusion terms selectively; avoid repeating long negative suffix bundles across the batch.
  \item If recent high-yield queries are topically narrow, expand laterally to adjacent entities/subdomains instead of refining the same pattern.
  \item Bias toward unexplored domains/entities/subtopics suggested by recent outcomes.
  \item Reduce patterns associated with low-yield patches.
  \item Prefer precise, verifiable source discovery with HTML-first pages.
  \item Avoid filetype filters by default; do not use \texttt{filetype:pdf} unless absolutely necessary.
  \item Do not output explanations outside JSON.
\end{enumerate}

\medskip
\noindent Return JSON with exactly this schema:
\begin{verbatim}
{
  "queries": ["q1", "q2"],
  "reason": "short rationale tied to recent outcomes"
}
\end{verbatim}
\end{promptbox}

\begin{promptbox}{Scout No-Fresh-Results Retry Instruction}
Retry condition: The previous discovery queries returned no fresh pages after filtering. This is retry attempt \texttt{<retry\_attempt>}. Generate substantially more diverse query structures and entity angles than before. Do not reuse operator skeletons from the previous batch. Avoid near-duplicates of prior phrasing and operator patterns. If recent high-yield queries are topically narrow, expand laterally to adjacent entities/subdomains instead of refining the same pattern. Prioritize novelty while staying on-topic.
\end{promptbox}

\begin{promptbox}{Scout Recent Query Outcomes Section}
The adaptive Scout re-seed prompt includes recent query outcomes from the run state. The section contains up to \texttt{<query\_history\_limit>} recent queries, with a default limit of 12. Each row is serialized as:
\begin{verbatim}
- query: "<query>" | relevant pages: <count> | gain-cost ratio: <ratio>
\end{verbatim}

\noindent If a recent query has no associated patch record, the row is serialized as:
\begin{verbatim}
- query: "<query_text>" | relevant pages: n/a | gain-cost ratio: n/a
\end{verbatim}

\noindent If there is no query history, the section is:
\begin{verbatim}
- <no prior queries>
\end{verbatim}
\end{promptbox}

\subsection{Forager Agent}

\begin{promptbox}{Forager Link-Scoring System Prompt}
You are the Forager Agent in a thematic web data collection system. You score candidate links for traversal priority. Precision is more important than recall. Return only valid JSON.
\end{promptbox}

\begin{promptbox}{Forager Link-Scoring Prompt}
Task: Score candidate links for thematic relevance and expected utility for local traversal.

\medskip
\noindent Theme description:
\texttt{<theme\_description>}

\medskip
\noindent \texttt{<topic-specific rules>}

\medskip
\noindent \texttt{<recent scored-link outcomes for calibration>}

\medskip
\noindent Candidate links JSON:
\begin{verbatim}
{
  "theme_description": "<theme_description>",
  "candidate_links": [
    {
      "url": "...",
      "anchor_text": "...",
      "surrounding_text": "...",
      "parent_url": "...",
      "source_url": "...",
      "source_title": "...",
      "source_snippet": "..."
    }
  ]
}
\end{verbatim}

\medskip
\noindent Scoring guidelines:
\begin{itemize}[leftmargin=1.5em, itemsep=0pt]
  \item Score each URL in $[0, 1]$.
  \item Use URL, anchor text, surrounding text, source title, and source snippet jointly.
  \item Prefer links likely to contain substantive thematic content.
  \item Penalize generic navigation/account/help/legal/cart/login pages, weak-context links, or likely low-information targets.
  \item If uncertain, score lower.
\end{itemize}

\medskip
\noindent Use exactly these four score bins:
\begin{enumerate}[leftmargin=1.5em, itemsep=0pt]
  \item $0.85$ to $1.00$: Highly likely relevant and high utility for patch exploitation.
  \item $0.60$ to $0.84$: Likely relevant, moderate-to-high utility.
  \item $0.30$ to $0.59$: Weak/indirect relevance, low utility.
  \item $0.00$ to $0.29$: Likely irrelevant or unproductive.
\end{enumerate}

\medskip
\noindent Output constraints:
\begin{enumerate}[leftmargin=1.5em, itemsep=0pt]
  \item Score every input URL exactly once.
  \item Do not invent or omit URLs.
  \item Do not include explanations or reasons.
  \item Return JSON only.
\end{enumerate}

\medskip
\noindent Return JSON with exactly this schema:
\begin{verbatim}
{
  "scored_links": [
    {"url": "...", "score": 0.0}
  ]
}
\end{verbatim}
\end{promptbox}

\begin{promptbox}{Forager Feedback Calibration Section}
Recent scored-link outcomes for calibration:
These are examples of what was previously sent to the Forager scorer and what happened after Critic classification.
\begin{verbatim}
{
  "positive_examples": [
    {
      "url": "...",
      "anchor_text": "...",
      "surrounding_text": "...",
      "source_title": "...",
      "source_snippet": "...",
      "assigned_score": 0.0,
      "critic_relevance": true,
      "critic_reason": "..."
    }
  ],
  "negative_examples": [...]
}
\end{verbatim}
\end{promptbox}

\subsection{Critic Agent}

\begin{promptbox}{Critic System Prompt}
You are the Critic Agent for binary relevance classification in thematic web-data collection. Classify each page as relevant or not relevant to the theme. Return only valid JSON.
\end{promptbox}

\begin{promptbox}{Critic Batch Prompt}
Task: Theme classification for multiple pages. Return JSON only.

\medskip
\noindent Theme description:
\texttt{<theme\_description>}

\medskip
\noindent \texttt{<topic-specific rules>}

\medskip
\noindent Decision rule:
\begin{itemize}[leftmargin=1.5em, itemsep=0pt]
  \item ``Yes'' if directly relevant to the theme or a strong bridge page.
  \item ``No'' if weak relevance, generic/navigation/legal-only, or low-information.
\end{itemize}

\medskip
\noindent Input JSON:
\begin{verbatim}
{
  "theme_description": "<theme_description>",
  "pages": [
    {"id": 0, "title": "...", "snippet": "..."}
  ]
}
\end{verbatim}

\noindent Output constraints:
\begin{enumerate}[leftmargin=1.5em, itemsep=0pt]
  \item Return exactly one result for each input id.
  \item Do not invent or omit ids.
  \item Confidence in $[0, 1]$.
  \item Reason $\leq 40$ words.
\end{enumerate}

\medskip
\noindent Return JSON schema:
\begin{verbatim}
{
  "results": [
    {"id": 0, "pred": "Yes" or "No", "reason": "...", "confidence": 0.0}
  ]
}
\end{verbatim}
\end{promptbox}

\subsection{Theme-Specific Prompt Elements}

The following \textbf{theme descriptions} were used for each theme, across all configurations:

\begin{itemize}[nosep]
    \item University: ``Four year colleges or universities in the United States''
    \item Fortune 500: ``Fortune 500 company job postings, career pages, and employment opportunity pages that specifically describe jobs or hiring opportunities at Fortune 500 companies''
    \item Law Schools: ``Accredited law schools in the United States (ABA-approved)''
    \item University: ``Autism support programs at colleges and universities in the United States that provide additional support beyond standard disability services specifically for autistic students''
    
\end{itemize}

The following addenda are inserted conditionally based on the configured theme. Each theme contributes separate instructions to the Scout, Forager, and Critic prompts.

\begin{promptbox}{Topic Addendum: U.S. Four-Year Colleges}
\textbf{Scout guidance.}
Prioritize official U.S. four-year institution pages; use negatives for community-college, two-year, law, medical, and graduate-only drift; avoid alumni, athletics, giving, bookstore, events, login, portal, and help surfaces.

\medskip
\noindent\textbf{Forager guidance.}
Score official institution home, about, admissions, undergraduate, and academics pages highest; score auxiliary institution pages lower; penalize two-year, professional-only, alumni, athletics, giving, events, login, and help pages.

\medskip
\noindent\textbf{Critic rule.}
Classify pages hosted by qualifying U.S. four-year college or university domains as relevant, including auxiliary pages; reject two-year, standalone professional/graduate-only, and unrelated third-party sites.
\end{promptbox}

\begin{promptbox}{Topic Addendum: U.S. Accredited Law Schools}
\textbf{Scout guidance.}
Prioritize official U.S. law school pages and authoritative accreditation/listing sources; favor ABA-approved schools, J.D. admissions, and school-of-law domains; avoid directories, forums, generic legal news, bar prep, law firms, and legal aid.

\medskip
\noindent\textbf{Forager guidance.}
Score official U.S. law school, ABA/accreditation, J.D. admissions, curriculum, tuition, faculty, and accreditation pages highest; score clear university routes to law schools moderately; penalize unrelated legal content and boilerplate.

\medskip
\noindent\textbf{Critic rule.}
Classify official pages for U.S. accredited/ABA-approved law schools or their university law-school sections as relevant; reject law firms, bar prep, legal aid, legal news/blog pages, and non-U.S. or non-accredited institutions.
\end{promptbox}

\begin{promptbox}{Topic Addendum: College Autism Support Programs}
\textbf{Scout guidance.}
Prioritize official college/university autism, ASD, and neurodiversity support program pages; require support beyond ordinary accommodations; avoid K--12, clinics, therapy providers, research-only centers, awareness pages, and parent-resource directories.

\medskip
\noindent\textbf{Forager guidance.}
Score official autism, ASD, neurodiversity support program pages and program admissions/services/coaching/transition/mentoring/eligibility/FAQ pages highest; penalize generic accommodations, K--12, clinic, research-only, awareness, parent-resource, login/help, and directory-only pages.

\medskip
\noindent\textbf{Critic rule.}
Classify pages describing autism-specific, ASD-specific, or neurodiversity support programs for college students as relevant, including pages listing such programs; reject generic disability services unless autism-specific structured programming is clear.
\end{promptbox}

\begin{promptbox}{Topic Addendum: Fortune 500 Job Postings}
\textbf{Scout guidance.}
Prioritize official company career sites, job-detail pages, and ATS surfaces with clear employer and opportunity signals; avoid news, investor pages, aggregators, salary/review sites, recruiting agencies, and generic career advice.

\medskip
\noindent\textbf{Forager guidance.}
Score official job-detail, career-search, application, ATS, internship, early-career, team, location, and hiring-program pages highest; score career landings moderately; penalize culture-only pages, aggregators, recruiters, news, investor, press, advice, login/help, and product pages.

\medskip
\noindent\textbf{Critic rule.}
Classify official company career pages or ATS-hosted pages describing employment opportunities at Fortune 500 companies as relevant; reject aggregators, salary/review pages, recruiters, advice, investor/news/press pages, and pages without employment/application content.
\end{promptbox}


\newpage
\section*{NeurIPS Paper Checklist}

\begin{enumerate}

\item {\bf Claims}
    \item[] Question: Do the main claims made in the abstract and introduction accurately reflect the paper's contributions and scope?
    \item[] Answer: \answerYes{}
    \item[] Justification: The claims made in the abstract and introduction accurately reflect the results reported in Section~\ref{results} based on the experiments detailed in Section~\ref{exp_setup}.
    \item[] Guidelines:
    \begin{itemize}
        \item The answer \answerNA{} means that the abstract and introduction do not include the claims made in the paper.
        \item The abstract and/or introduction should clearly state the claims made, including the contributions made in the paper and important assumptions and limitations. A \answerNo{} or \answerNA{} answer to this question will not be perceived well by the reviewers. 
        \item The claims made should match theoretical and experimental results, and reflect how much the results can be expected to generalize to other settings. 
        \item It is fine to include aspirational goals as motivation as long as it is clear that these goals are not attained by the paper. 
    \end{itemize}

\item {\bf Limitations}
    \item[] Question: Does the paper discuss the limitations of the work performed by the authors?
    \item[] Answer: \answerYes{}
    \item[] Justification: Limitations introduced by the system design and experimental setting are discussed in Section~\ref{limitations}.
    \item[] Guidelines:
    \begin{itemize}
        \item The answer \answerNA{} means that the paper has no limitation while the answer \answerNo{} means that the paper has limitations, but those are not discussed in the paper. 
        \item The authors are encouraged to create a separate ``Limitations'' section in their paper.
        \item The paper should point out any strong assumptions and how robust the results are to violations of these assumptions (e.g., independence assumptions, noiseless settings, model well-specification, asymptotic approximations only holding locally). The authors should reflect on how these assumptions might be violated in practice and what the implications would be.
        \item The authors should reflect on the scope of the claims made, e.g., if the approach was only tested on a few datasets or with a few runs. In general, empirical results often depend on implicit assumptions, which should be articulated.
        \item The authors should reflect on the factors that influence the performance of the approach. For example, a facial recognition algorithm may perform poorly when image resolution is low or images are taken in low lighting. Or a speech-to-text system might not be used reliably to provide closed captions for online lectures because it fails to handle technical jargon.
        \item The authors should discuss the computational efficiency of the proposed algorithms and how they scale with dataset size.
        \item If applicable, the authors should discuss possible limitations of their approach to address problems of privacy and fairness.
        \item While the authors might fear that complete honesty about limitations might be used by reviewers as grounds for rejection, a worse outcome might be that reviewers discover limitations that aren't acknowledged in the paper. The authors should use their best judgment and recognize that individual actions in favor of transparency play an important role in developing norms that preserve the integrity of the community. Reviewers will be specifically instructed to not penalize honesty concerning limitations.
    \end{itemize}

\item {\bf Theory assumptions and proofs}
    \item[] Question: For each theoretical result, does the paper provide the full set of assumptions and a complete (and correct) proof?
    \item[] Answer: \answerNA{}
    \item[] Justification: We provide theoretical framing of the problem in Section~\ref{prob_def} with a full set of assumptions, but do not claim to make any theoretical result or proof.
    \item[] Guidelines:
    \begin{itemize}
        \item The answer \answerNA{} means that the paper does not include theoretical results. 
        \item All the theorems, formulas, and proofs in the paper should be numbered and cross-referenced.
        \item All assumptions should be clearly stated or referenced in the statement of any theorems.
        \item The proofs can either appear in the main paper or the supplemental material, but if they appear in the supplemental material, the authors are encouraged to provide a short proof sketch to provide intuition. 
        \item Inversely, any informal proof provided in the core of the paper should be complemented by formal proofs provided in appendix or supplemental material.
        \item Theorems and Lemmas that the proof relies upon should be properly referenced. 
    \end{itemize}

    \item {\bf Experimental result reproducibility}
    \item[] Question: Does the paper fully disclose all the information needed to reproduce the main experimental results of the paper to the extent that it affects the main claims and/or conclusions of the paper (regardless of whether the code and data are provided or not)?
    \item[] Answer: \answerYes{}
    \item[] Justification: The framework and experiments are described in sufficient detail for reproducibility of the experiments in Sections~\ref{method} and ~\ref{exp_setup} and Appendix~\ref{prompts}. As mentioned in Section~\ref{limitations}, experiments were conducted on the live web, so exact reproducibility of the reported results is limited.
    \item[] Guidelines:
    \begin{itemize}
        \item The answer \answerNA{} means that the paper does not include experiments.
        \item If the paper includes experiments, a \answerNo{} answer to this question will not be perceived well by the reviewers: Making the paper reproducible is important, regardless of whether the code and data are provided or not.
        \item If the contribution is a dataset and\slash or model, the authors should describe the steps taken to make their results reproducible or verifiable. 
        \item Depending on the contribution, reproducibility can be accomplished in various ways. For example, if the contribution is a novel architecture, describing the architecture fully might suffice, or if the contribution is a specific model and empirical evaluation, it may be necessary to either make it possible for others to replicate the model with the same dataset, or provide access to the model. In general. releasing code and data is often one good way to accomplish this, but reproducibility can also be provided via detailed instructions for how to replicate the results, access to a hosted model (e.g., in the case of a large language model), releasing of a model checkpoint, or other means that are appropriate to the research performed.
        \item While NeurIPS does not require releasing code, the conference does require all submissions to provide some reasonable avenue for reproducibility, which may depend on the nature of the contribution. For example
        \begin{enumerate}
            \item If the contribution is primarily a new algorithm, the paper should make it clear how to reproduce that algorithm.
            \item If the contribution is primarily a new model architecture, the paper should describe the architecture clearly and fully.
            \item If the contribution is a new model (e.g., a large language model), then there should either be a way to access this model for reproducing the results or a way to reproduce the model (e.g., with an open-source dataset or instructions for how to construct the dataset).
            \item We recognize that reproducibility may be tricky in some cases, in which case authors are welcome to describe the particular way they provide for reproducibility. In the case of closed-source models, it may be that access to the model is limited in some way (e.g., to registered users), but it should be possible for other researchers to have some path to reproducing or verifying the results.
        \end{enumerate}
    \end{itemize}

\item {\bf Open access to data and code}
    \item[] Question: Does the paper provide open access to the data and code, with sufficient instructions to faithfully reproduce the main experimental results, as described in supplemental material?
    \item[] Answer: \answerYes{}
    \item[] Justification: We will release code and specific instructions reproducibility for the paper upon acceptance, or during the review process if requested.
    \item[] Guidelines:
    \begin{itemize}
        \item The answer \answerNA{} means that paper does not include experiments requiring code.
        \item Please see the NeurIPS code and data submission guidelines (\url{https://neurips.cc/public/guides/CodeSubmissionPolicy}) for more details.
        \item While we encourage the release of code and data, we understand that this might not be possible, so \answerNo{} is an acceptable answer. Papers cannot be rejected simply for not including code, unless this is central to the contribution (e.g., for a new open-source benchmark).
        \item The instructions should contain the exact command and environment needed to run to reproduce the results. See the NeurIPS code and data submission guidelines (\url{https://neurips.cc/public/guides/CodeSubmissionPolicy}) for more details.
        \item The authors should provide instructions on data access and preparation, including how to access the raw data, preprocessed data, intermediate data, and generated data, etc.
        \item The authors should provide scripts to reproduce all experimental results for the new proposed method and baselines. If only a subset of experiments are reproducible, they should state which ones are omitted from the script and why.
        \item At submission time, to preserve anonymity, the authors should release anonymized versions (if applicable).
        \item Providing as much information as possible in supplemental material (appended to the paper) is recommended, but including URLs to data and code is permitted.
    \end{itemize}

\item {\bf Experimental setting/details}
    \item[] Question: Does the paper specify all the training and test details (e.g., data splits, hyperparameters, how they were chosen, type of optimizer) necessary to understand the results?
    \item[] Answer: \answerYes{}
    \item[] Justification: Section~\ref{exp_setup} describes the experimental setting in sufficient detail to understand the results.
    \item[] Guidelines:
    \begin{itemize}
        \item The answer \answerNA{} means that the paper does not include experiments.
        \item The experimental setting should be presented in the core of the paper to a level of detail that is necessary to appreciate the results and make sense of them.
        \item The full details can be provided either with the code, in appendix, or as supplemental material.
    \end{itemize}

\item {\bf Experiment statistical significance}
    \item[] Question: Does the paper report error bars suitably and correctly defined or other appropriate information about the statistical significance of the experiments?
    \item[] Answer: \answerNo
    \item[] Justification: Each system configuration is evaluated in a single run due to the high computational cost of collection (approximately 3 hours and 1.1M LLM tokens per run). Running multiple trials per configuration was not feasible within the available budget. We report endpoint metrics and coverage curves across four distinct topics to provide breadth of evidence.
    \item[] Guidelines:
    \begin{itemize}
        \item The answer \answerNA{} means that the paper does not include experiments.
        \item The authors should answer \answerYes{} if the results are accompanied by error bars, confidence intervals, or statistical significance tests, at least for the experiments that support the main claims of the paper.
        \item The factors of variability that the error bars are capturing should be clearly stated (for example, train/test split, initialization, random drawing of some parameter, or overall run with given experimental conditions).
        \item The method for calculating the error bars should be explained (closed form formula, call to a library function, bootstrap, etc.)
        \item The assumptions made should be given (e.g., Normally distributed errors).
        \item It should be clear whether the error bar is the standard deviation or the standard error of the mean.
        \item It is OK to report 1-sigma error bars, but one should state it. The authors should preferably report a 2-sigma error bar than state that they have a 96\% CI, if the hypothesis of Normality of errors is not verified.
        \item For asymmetric distributions, the authors should be careful not to show in tables or figures symmetric error bars that would yield results that are out of range (e.g., negative error rates).
        \item If error bars are reported in tables or plots, the authors should explain in the text how they were calculated and reference the corresponding figures or tables in the text.
    \end{itemize}

\item {\bf Experiments compute resources}
    \item[] Question: For each experiment, does the paper provide sufficient information on the computer resources (type of compute workers, memory, time of execution) needed to reproduce the experiments?
    \item[] Answer: \answerYes{}
    \item[] Justification: Appendix~\ref{cost-analysis} gives a full account of the resources utilized to execute experiments.
    \item[] Guidelines:
    \begin{itemize}
        \item The answer \answerNA{} means that the paper does not include experiments.
        \item The paper should indicate the type of compute workers CPU or GPU, internal cluster, or cloud provider, including relevant memory and storage.
        \item The paper should provide the amount of compute required for each of the individual experimental runs as well as estimate the total compute. 
        \item The paper should disclose whether the full research project required more compute than the experiments reported in the paper (e.g., preliminary or failed experiments that didn't make it into the paper). 
    \end{itemize}
    
\item {\bf Code of ethics}
    \item[] Question: Does the research conducted in the paper conform, in every respect, with the NeurIPS Code of Ethics \url{https://neurips.cc/public/EthicsGuidelines}?
    \item[] Answer: \answerYes{}
    \item[] Justification: The research involved in this work conforms to the Code of Ethics.
    \item[] Guidelines:
    \begin{itemize}
        \item The answer \answerNA{} means that the authors have not reviewed the NeurIPS Code of Ethics.
        \item If the authors answer \answerNo, they should explain the special circumstances that require a deviation from the Code of Ethics.
        \item The authors should make sure to preserve anonymity (e.g., if there is a special consideration due to laws or regulations in their jurisdiction).
    \end{itemize}

\item {\bf Broader impacts}
    \item[] Question: Does the paper discuss both potential positive societal impacts and negative societal impacts of the work performed?
    \item[] Answer: \answerYes{}
    \item[] Justification: We note that the reported work has the potential to benefit social science researchers. We acknowledge that comprehensive thematic web data collection is dual-use and carries inherent risk for surveillance and malicious scraping. We mitigate this risk by implementing the system with a read-only design, and do not release any model or corpus. Additionally, we implement standard practices of `crawler etiquette'; our system reads and respects robots.txt where available, enforces a global concurrency cap of eight requests, does not revisit a URL within a run, and does not retry requests returning access-control or rate-limit responses such as HTTP 403 or 429. This was implemented throughout all reported experiments.
    \item[] Guidelines:
    \begin{itemize}
        \item The answer \answerNA{} means that there is no societal impact of the work performed.
        \item If the authors answer \answerNA{} or \answerNo, they should explain why their work has no societal impact or why the paper does not address societal impact.
        \item Examples of negative societal impacts include potential malicious or unintended uses (e.g., disinformation, generating fake profiles, surveillance), fairness considerations (e.g., deployment of technologies that could make decisions that unfairly impact specific groups), privacy considerations, and security considerations.
        \item The conference expects that many papers will be foundational research and not tied to particular applications, let alone deployments. However, if there is a direct path to any negative applications, the authors should point it out. For example, it is legitimate to point out that an improvement in the quality of generative models could be used to generate Deepfakes for disinformation. On the other hand, it is not needed to point out that a generic algorithm for optimizing neural networks could enable people to train models that generate Deepfakes faster.
        \item The authors should consider possible harms that could arise when the technology is being used as intended and functioning correctly, harms that could arise when the technology is being used as intended but gives incorrect results, and harms following from (intentional or unintentional) misuse of the technology.
        \item If there are negative societal impacts, the authors could also discuss possible mitigation strategies (e.g., gated release of models, providing defenses in addition to attacks, mechanisms for monitoring misuse, mechanisms to monitor how a system learns from feedback over time, improving the efficiency and accessibility of ML).
    \end{itemize}
    
\item {\bf Safeguards}
    \item[] Question: Does the paper describe safeguards that have been put in place for responsible release of data or models that have a high risk for misuse (e.g., pre-trained language models, image generators, or scraped datasets)?
    \item[] Answer: \answerNA{}
    \item[] Justification: We do not release any dataset or models.
    \item[] Guidelines:
    \begin{itemize}
        \item The answer \answerNA{} means that the paper poses no such risks.
        \item Released models that have a high risk for misuse or dual-use should be released with necessary safeguards to allow for controlled use of the model, for example by requiring that users adhere to usage guidelines or restrictions to access the model or implementing safety filters. 
        \item Datasets that have been scraped from the Internet could pose safety risks. The authors should describe how they avoided releasing unsafe images.
        \item We recognize that providing effective safeguards is challenging, and many papers do not require this, but we encourage authors to take this into account and make a best faith effort.
    \end{itemize}

\item {\bf Licenses for existing assets}
    \item[] Question: Are the creators or original owners of assets (e.g., code, data, models), used in the paper, properly credited and are the license and terms of use explicitly mentioned and properly respected?
    \item[] Answer: \answerYes{}
    \item[] Justification: The only existing assest used, DeepResearcher, is cited in the main text and the Apache 2.0 License it is released under is explicity mentioned in Appendix~\ref{deepresearcher-detail}.
    \item[] Guidelines:
    \begin{itemize}
        \item The answer \answerNA{} means that the paper does not use existing assets.
        \item The authors should cite the original paper that produced the code package or dataset.
        \item The authors should state which version of the asset is used and, if possible, include a URL.
        \item The name of the license (e.g., CC-BY 4.0) should be included for each asset.
        \item For scraped data from a particular source (e.g., website), the copyright and terms of service of that source should be provided.
        \item If assets are released, the license, copyright information, and terms of use in the package should be provided. For popular datasets, \url{paperswithcode.com/datasets} has curated licenses for some datasets. Their licensing guide can help determine the license of a dataset.
        \item For existing datasets that are re-packaged, both the original license and the license of the derived asset (if it has changed) should be provided.
        \item If this information is not available online, the authors are encouraged to reach out to the asset's creators.
    \end{itemize}

\item {\bf New assets}
    \item[] Question: Are new assets introduced in the paper well documented and is the documentation provided alongside the assets?
    \item[] Answer: \answerNA{}
    \item[] Justification: We do not release new assets.
    \item[] Guidelines:
    \begin{itemize}
        \item The answer \answerNA{} means that the paper does not release new assets.
        \item Researchers should communicate the details of the dataset\slash code\slash model as part of their submissions via structured templates. This includes details about training, license, limitations, etc. 
        \item The paper should discuss whether and how consent was obtained from people whose asset is used.
        \item At submission time, remember to anonymize your assets (if applicable). You can either create an anonymized URL or include an anonymized zip file.
    \end{itemize}

\item {\bf Crowdsourcing and research with human subjects}
    \item[] Question: For crowdsourcing experiments and research with human subjects, does the paper include the full text of instructions given to participants and screenshots, if applicable, as well as details about compensation (if any)? 
    \item[] Answer: \answerNA{}
    \item[] Justification: The paper does not  involve crowdsourcing nor research with human subjects.
    \item[] Guidelines:
    \begin{itemize}
        \item The answer \answerNA{} means that the paper does not involve crowdsourcing nor research with human subjects.
        \item Including this information in the supplemental material is fine, but if the main contribution of the paper involves human subjects, then as much detail as possible should be included in the main paper. 
        \item According to the NeurIPS Code of Ethics, workers involved in data collection, curation, or other labor should be paid at least the minimum wage in the country of the data collector. 
    \end{itemize}

\item {\bf Institutional review board (IRB) approvals or equivalent for research with human subjects}
    \item[] Question: Does the paper describe potential risks incurred by study participants, whether such risks were disclosed to the subjects, and whether Institutional Review Board (IRB) approvals (or an equivalent approval/review based on the requirements of your country or institution) were obtained?
    \item[] Answer: \answerNA{}
    \item[] Justification: The paper does not involve crowdsourcing nor research with human subjects.
    \item[] Guidelines:
    \begin{itemize}
        \item The answer \answerNA{} means that the paper does not involve crowdsourcing nor research with human subjects.
        \item Depending on the country in which research is conducted, IRB approval (or equivalent) may be required for any human subjects research. If you obtained IRB approval, you should clearly state this in the paper. 
        \item We recognize that the procedures for this may vary significantly between institutions and locations, and we expect authors to adhere to the NeurIPS Code of Ethics and the guidelines for their institution. 
        \item For initial submissions, do not include any information that would break anonymity (if applicable), such as the institution conducting the review.
    \end{itemize}

\item {\bf Declaration of LLM usage}
    \item[] Question: Does the paper describe the usage of LLMs if it is an important, original, or non-standard component of the core methods in this research? Note that if the LLM is used only for writing, editing, or formatting purposes and does \emph{not} impact the core methodology, scientific rigor, or originality of the research, declaration is not required.
    \item[] Answer: \answerYes{}
    \item[] Justification: The paper describes the use of LLMs in detail in Sections~\ref{method} and ~\ref{exp_setup}. Full prompts are given in Appendix~\ref{prompts}.
    \item[] Guidelines:
    \begin{itemize}
        \item The answer \answerNA{} means that the core method development in this research does not involve LLMs as any important, original, or non-standard components.
        \item Please refer to our LLM policy in the NeurIPS handbook for what should or should not be described.
    \end{itemize}

\end{enumerate}

\end{document}